%% file: main.tex
\documentclass[10pt,twocolumn,letterpaper]{article}

\usepackage{cvpr}              %

\input{preamble}

\definecolor{cvprblue}{rgb}{0.21,0.49,0.74}
\usepackage[pagebackref,breaklinks,colorlinks,citecolor=cvprblue]{hyperref}

\def\paperID{469} %
\def\confName{CVPR}
\def\confYear{2024}

\newcommand{\projectname}[0]{OcMesher}

\title{View-Dependent Octree-based Mesh Extraction in Unbounded Scenes for Procedural Synthetic Data}

\author{Zeyu Ma \quad Alexander Raistrick \quad Lahav Lipson \quad Jia Deng \\
Department of Computer Science, Princeton University\\
{\tt\small \{zeyum, araistrick, llipson, jiadeng\}@princeton.edu}
}

\begin{document}
\maketitle
\begin{abstract}

Procedural synthetic data generation has received increasing attention in computer vision. Procedural signed distance functions (SDFs) are a powerful tool for modeling large-scale detailed scenes, but existing mesh extraction methods have artifacts or performance profiles that limit their use for synthetic data. We propose \projectname{}, a mesh extraction algorithm that efficiently handles high-detail unbounded scenes with perfect view-consistency, with easy export to downstream real-time engines. The main novelty of our solution is an algorithm to construct an octree based on a given SDF and multiple camera views. We performed extensive experiments, and show our solution produces better synthetic data for training and evaluation of computer vision models. 

\end{abstract}

\vspace{-3mm}

\section{Introduction}
\label{sec:intro}

\begin{figure*}[t]
  \centering
    \includegraphics[width=\linewidth]{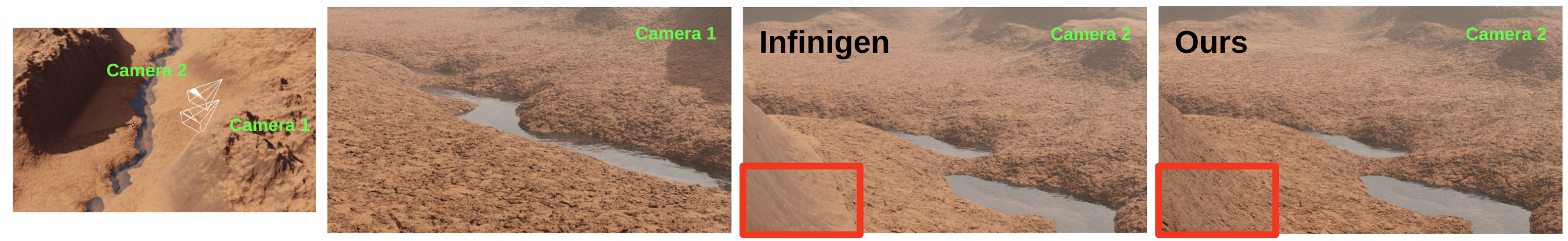}

   \caption{ The \infinigen{} solution has low poly artifacts if the camera changes its location significantly, while our solution has no artifacts.}
   \label{fig:scene_contrast}

\vspace{-4mm}

\end{figure*}

Synthetic visual data rendered from conventional computer graphics has seen increasing use in computer vision~\cite{gta5koltun2016playing, crestereo, law2022label, bai2022deep, lipson2022coupled, ma2022multiview, deitke2022procthor,raistrick2023infinite,jiang2018configurable,li2018interiornet,devaranjan2020meta,tsirikoglou2017procedural,wrenninge2018synscapes,khan2019procsy,greff2022kubric,yao2020blendedmvs,butler2012naturalistic}. In 3D vision and embodied AI, synthetic data has been frequently used for training~\cite{law2022label,yao2020blendedmvs,wang2020tartanair,gta5koltun2016playing,butler2012naturalistic}, evaluation~\cite{yao2020blendedmvs,wang2020tartanair,butler2012naturalistic}, and constructing virtual environments~\cite{deitke2022procthor,li2018interiornet}. One advantage of synthetic data is that they can be procedurally generated, that is, created algorithmically by compact randomized mathematical rules. Procedural generation can greatly improve the diversity of synthetic data through infinite random variations and is a promising research direction that has received increasing attention. Procedural generation has been employed in generating both indoor scenes~\cite{greff2022kubric,deitke2022procthor} as well as outdoor scenes~\cite{devaranjan2020meta,raistrick2023infinite,wrenninge2018synscapes,khan2019procsy}. 

Procedural signed distance functions (SDF) are a common technique for procedural scene generation. An SDF is a function that maps a 3D point to a real number that represents the signed distance to a 3D surface. A procedural SDF is an SDF expressed by compact mathematical rules. For example, a mountain range can be modelled as a procedural SDF using Perlin noise~\cite{perlin1985image}. Procedural SDFs are powerful because just a few lines of code using primitive math functions can represent an arbitrarily large surface with arbitrarily fine details. For example, Infinigen~\cite{raistrick2023infinite}, a recently released procedural generator of 3D scenes, makes extensive use of procedural SDFs to generate varied and detailed terrains.  

To be used for synthetic data generation, an SDF must first be converted to a mesh, which is the universally supported format for common rendering engines. This process is called ``mesh extraction''. The standard approach is to evaluate the SDF on a uniformly-spaced regular 3D grid and use the ``Marching Cubes'' algorithm~\cite{lorensen1998marching}, which extracts a mesh from the cells that straddle the zero crossings of the SDF (i.e.\@ iso-surface extraction). 

However, this standard approach does not work for unbounded scenes, which are scenes that contain visible geometry that is arbitrarily far away. For example, an ocean that extends to the horizon is unbounded. For such unbounded scenes, a uniformly spaced regular SDF grid will be prohibitively large to cover far-away geometry. An existing solution is to use a regular 3D grid that is spaced non-uniformly in a view-dependent way, with smaller spacing closer to the camera. This allows far-away geometry to be represented by large faces in the mesh, vastly improving efficiency.  This is the solution adopted by Infinigen~\cite{raistrick2023infinite}, which uses a view-dependent non-uniform spacing based on spherical coordinates. For convenience, we will refer to this solution as the ``Infinigen solution''. 

The Infinigen solution, however, still suffers from a major drawback in terms of view consistency. In this solution, the spacing of the grid is centered around a single camera location, with small cells near the camera and large cells from far away, such that the polygonal faces of the extracted mesh at different distances will project to about the same size in the pixel space. This works well if the camera stays in the same location, but if the camera changes its location significantly, visual artifacts will occur if we render the same mesh in the new camera view: coarse parts of the mesh can become newly visible or closer to the camera, creating an undesirable ``low-poly'' appearance as shown in Fig. \ref{fig:scene_contrast}. 

These artifacts can be reduced by extracting a new view-dependent mesh when the camera changes position, which is the fix adopted by Infinigen. But this introduces two new issues: (1) generating a new mesh for each new view is costly and impractical for real-time rendering, which is desirable for simulated environments for embodied AI; and (2) switching between meshes causes severe flickering artifacts, unless they are extracted at prohibitively high resolution. 

In this paper, we propose \projectname{}, a new mesh extraction solution for unbounded scenes that avoids the pitfalls of existing solutions. \projectname{} can generate a single mesh that represents unbounded geometry efficiently and can be rendered without artifacts across any pre-defined range of camera views. Our solution achieves a \emph{combination} of capabilities that was not possible with existing solutions:
\begin{itemize}
    \item \emph{Unbounded scenes: } Our solution can efficiently handle unbounded scenes that include arbitrarily distant geometry, without intractable memory requirements. This is not possible with the naive solution of mesh extraction from a uniformly spaced regular grid. 
    \item \emph{Real-time rendering: } Our solution generates a single mesh that can be directly used by a real-time rendering engine like Unreal Engine \cite{unrealengine}. The mesh renders well across a range of camera views, as long as the camera stays within an area predefined by the user. This is not possible with the Infinigen solution that uses a view-dependent non-uniformly-spaced regular grid. 
    \item \emph{View consistency: } Because only a single mesh is needed, our solution has perfect view consistency, with zero flickering artifacts across the rendered video frames, regardless of the poly count of the mesh. Our perfect view consistency improves data quality. This is not possible with the Infinigen solution, which is not able to completely eliminate flickering even with a very high poly count. 
\end{itemize}

\begin{figure}[t]
\vspace{-3mm}
  \centering
\begin{subfigure}[t]{.3\linewidth}\centering
    \includegraphics[width=\linewidth]{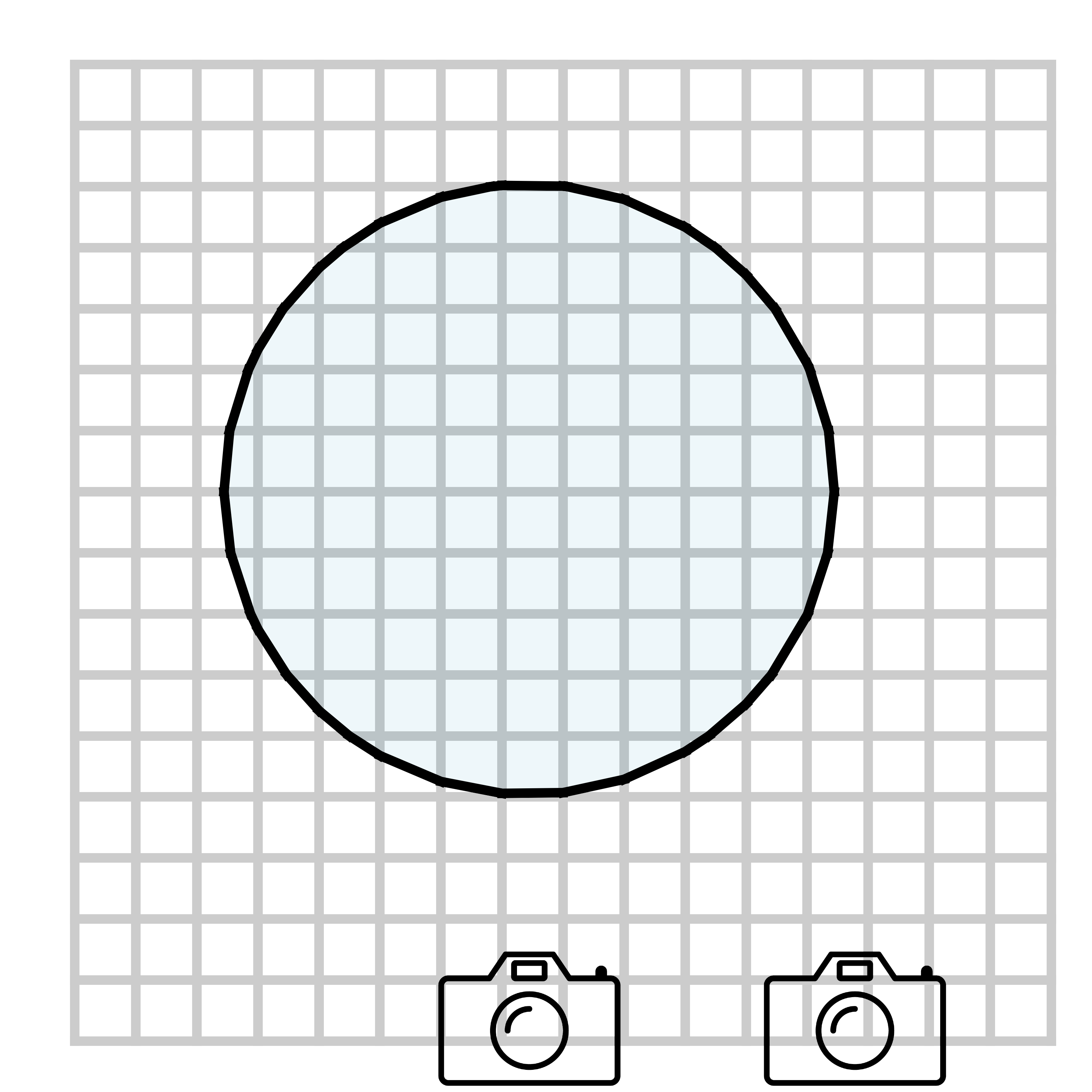}
    \caption*{(a)}
    \end{subfigure}
\begin{subfigure}[t]{.38\linewidth}\centering
    \includegraphics[width=\linewidth]{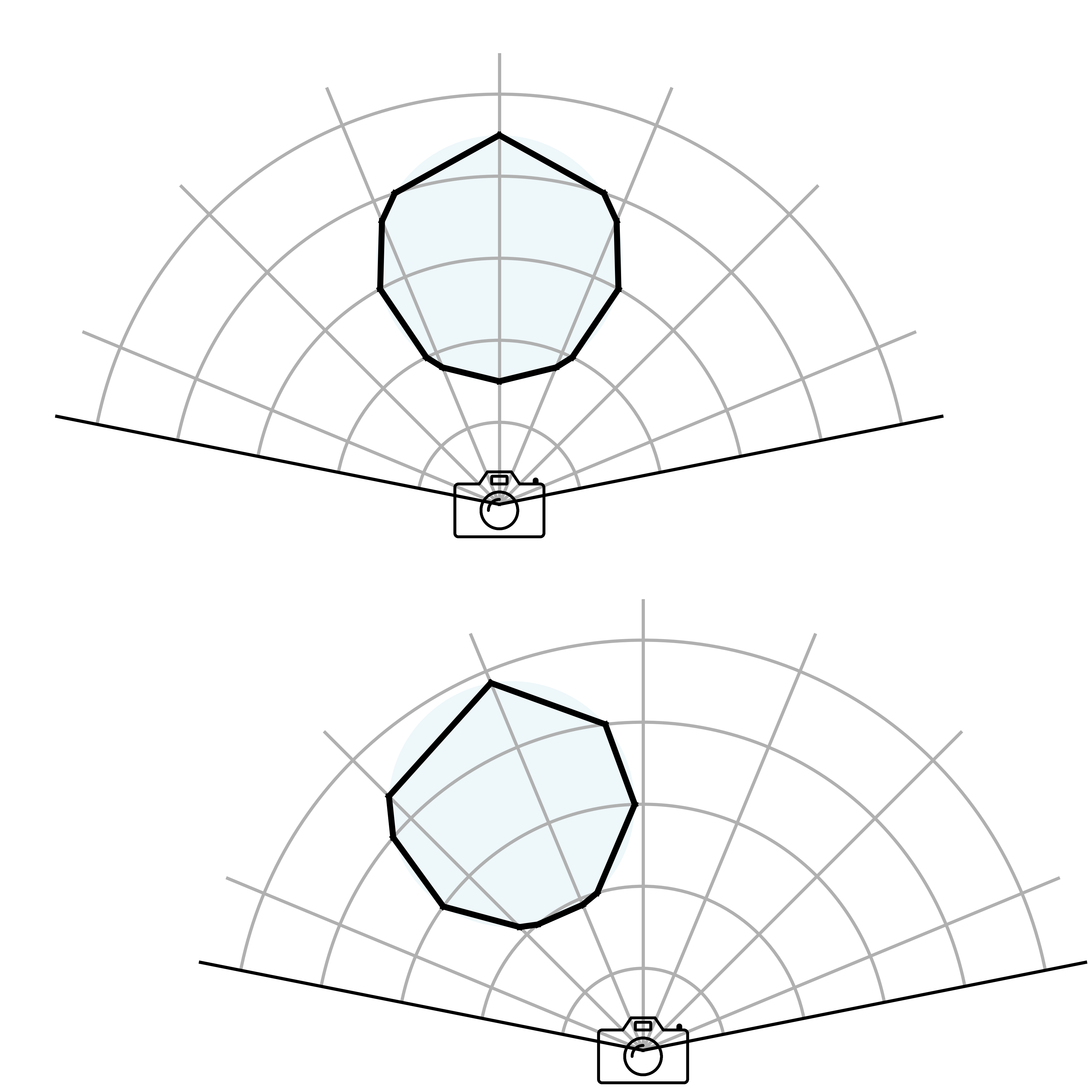}
    \caption*{(b)}
    \end{subfigure}
\begin{subfigure}[t]{.3\linewidth}\centering
    \includegraphics[width=\linewidth]{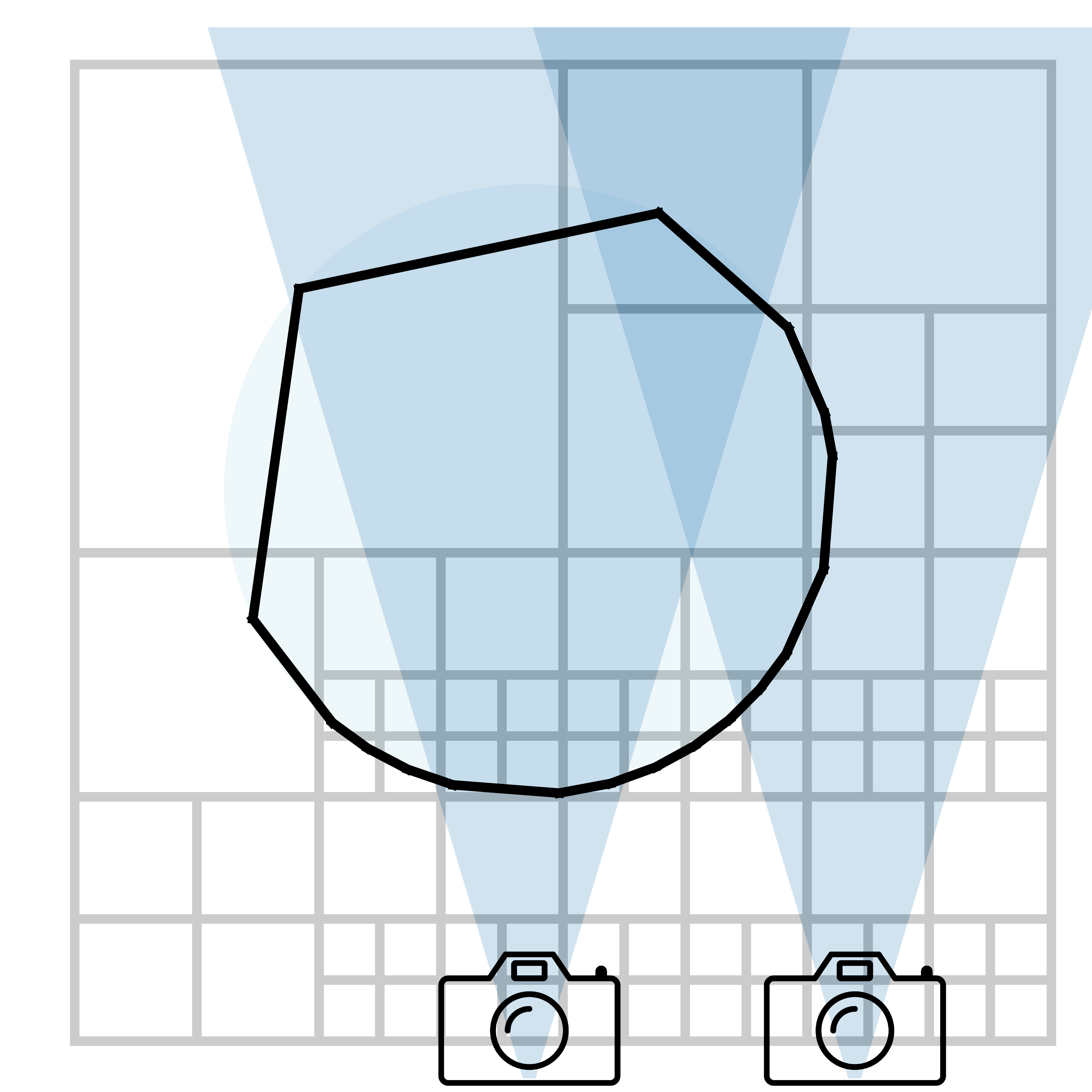}
    \caption*{(c)}
    \end{subfigure}

   \caption{(a) Uniform marching cubes mesh extraction from cannot efficiently represent high-detail unbounded scenes; (b) Infinigen's solution requires different meshes for new views, so introduces artifacts; (c) Our algorithm solves these issues by constructing an \emph{efficient} octree and extracts the mesh from it to solve the issues.}
   \label{fig:compare}
   \vspace{-5mm}
\end{figure}

The core idea of our solution is to construct an \emph{octree} \cite{octree} instead of a regular grid for mesh extraction. An octree can be understood as an irregular, multi-resolution grid. Given a user-defined range of camera views and a procedural SDF, we construct an octree that is high resolution around surfaces close to any camera view, but low resolution for locations in empty space or far away from all camera views. Our algorithm seeks to construct an \emph{efficient} octree, minimizing the number of total cells used while maintaining visual quality of the renders. Once the octree is constructed, we perform dual contouring~\cite{ju2002dual} to extract a mesh. Fig.~\ref{fig:compare} illustrates the idea in comparison with existing solutions. 

The main novelty of our solution lies in the algorithm that constructs an octree based on a given SDF and \emph{multiple} camera views. Existing techniques in the literature, including the computer graphics literature, have only addressed the octree construction dependent on a single view.  To the best of our knowledge, we are the first to construct octrees dependent on multiple views.

We perform extensive experiments to validate the effectiveness of our solution. Experiments show that our solution can indeed eliminate the flickering artifacts present in Infinigen's solution, leading to better synthetic data. In addition, we show our solution can generate unbounded environments that can be rendered at 50 FPS in Unreal Engine, seamlessly and without artifacts even with large view changes, a capability not available from existing solutions.
We encourage the reader to view the video \href{https://youtu.be/YA1c5L0Ncuw}{here} for more qualitative results and please visit \href{https://github.com/princeton-vl/OcMesher}{https://github.com/princeton-vl/OcMesher} for the code.

\section{Related Work}

Procedural signed distance functions have been widely used in realistic scene generation~\cite{musgrave1993methods,perlin2002hyper,geiss2007generating,raistrick2023infinite}, from height maps to increasingly complex 3D compositions. Approaches to visualize or render scenes modelled as SDFs fall into many categories.

\paragraph{SDF Rendering Algorithms}
A potential alternative to our approach is to directly render an SDF without first extracting a mesh, using algorithms such as sphere tracing~\cite{hart1996sphere}, segment tracing~\cite{galin2020segment}, quasi-analytic error-bounded (QAEB) ray tracing~\cite{musgrave2002qaeb}, or others~\cite{gobbetti2008single,crassin2009gigavoxels}, or even commercial engines \cite{VECTRON}. However, these generally require extra assumptions (e.g. Lipschitz continuity~\cite{galin2020segment}, or that the SDF represents a real Euclidean distance~\cite{hart1996sphere}), and are generally less performant and widely adopted than mesh-based renderers. Our mesh-extraction solution can be used in popular mesh-based engines and simulators (e.g. Blender \cite{blender}, Unreal Engine \cite{unrealengine}) without code additions, and trivially integrates with non-SDF-based assets (such as all assets besides terrain in Infinigen \cite{raistrick2023infinite}).

\paragraph{Multi-Resolution Mesh Extraction}
Uniform iso-surface extraction algorithms like Marching Cubes~\cite{lorensen1998marching} and Marching Tetrahedra \cite{marchingtetra} are the standard approach to mesh extraction for implicit functions. However, they are less well suited when scene content is unbounded or not all equally important. Therefore, many existing works divide space into multiresolution tetrahedra~\cite{zhou1997multiresolution,gerstner2000topology}, or even polyhedra~\cite{pascucci2002efficient,weber2003extraction}. Some works also divide space into an octree and use dual contouring algorithms~\cite{perry2001kizamu,ju2002dual,ju2006intersection}. Please refer to \cite{wenger2013isosurfaces} for a more complete discussion. However, these works do not answer the question of how to determine the level of detail (LOD) when we have multiple cameras.

\paragraph{View Dependent Mesh Extraction}
To the best of our knowledge, all existing works focus on a single camera at a time and reuse the mesh for very few neighboring cameras. Many construct volume grids in world space and subdivide them recursively~\cite{livnat1998view,liu2001progressive,gregorski2002interactive,scholz2015real}. These works usually do not optimize the mesh size based on criteria such as occlusion and do not scale well for high-resolution images. For example, Scholz et al.~\cite{scholz2015real} focuses on real-time approaches and their generated mesh contains at most 700k triangles for a $1920 \times 1080$ resolution image of more than 2 million pixels. Considering the deep depth dimension and the fact that they do not coarsen occluded triangles, their meshes are far from achieving pixel-level details in unbounded scenes.

Infinigen~\cite{raistrick2023infinite} uses Marching Cubes in camera-space spherical coordinates, which is more scalable to high-resolution images. However, as mentioned previously, it is impractical for real-time rendering and suffers flickering and low poly artifacts due to switching between view-dependent meshes as the camera moves.

\paragraph{Flickering Removal}
Besides generating a static mesh, alternative flickering removal approaches attempt to hide or smooth out LOD transitions by continuously deforming the mesh~\cite{hoppe1997view,lindstrom1996real}. However, these techniques usually require the geometry to be a height map~\cite{lindstrom1996real} or given as an initial high-poly mesh~\cite{hoppe1997view}, which is not applicable in our case with an SDF as input. Other blending algorithms only consider changes in image space \cite{giegl2007unpopping}, ignoring 3D geometry. These solutions also cannot export a static mesh for use in do real-time rendering engines.

\begin{figure}[b]
\vspace{-3mm}
  \centering
\begin{subfigure}[t]{.48\linewidth}\centering
    \includegraphics[width=.67\linewidth]{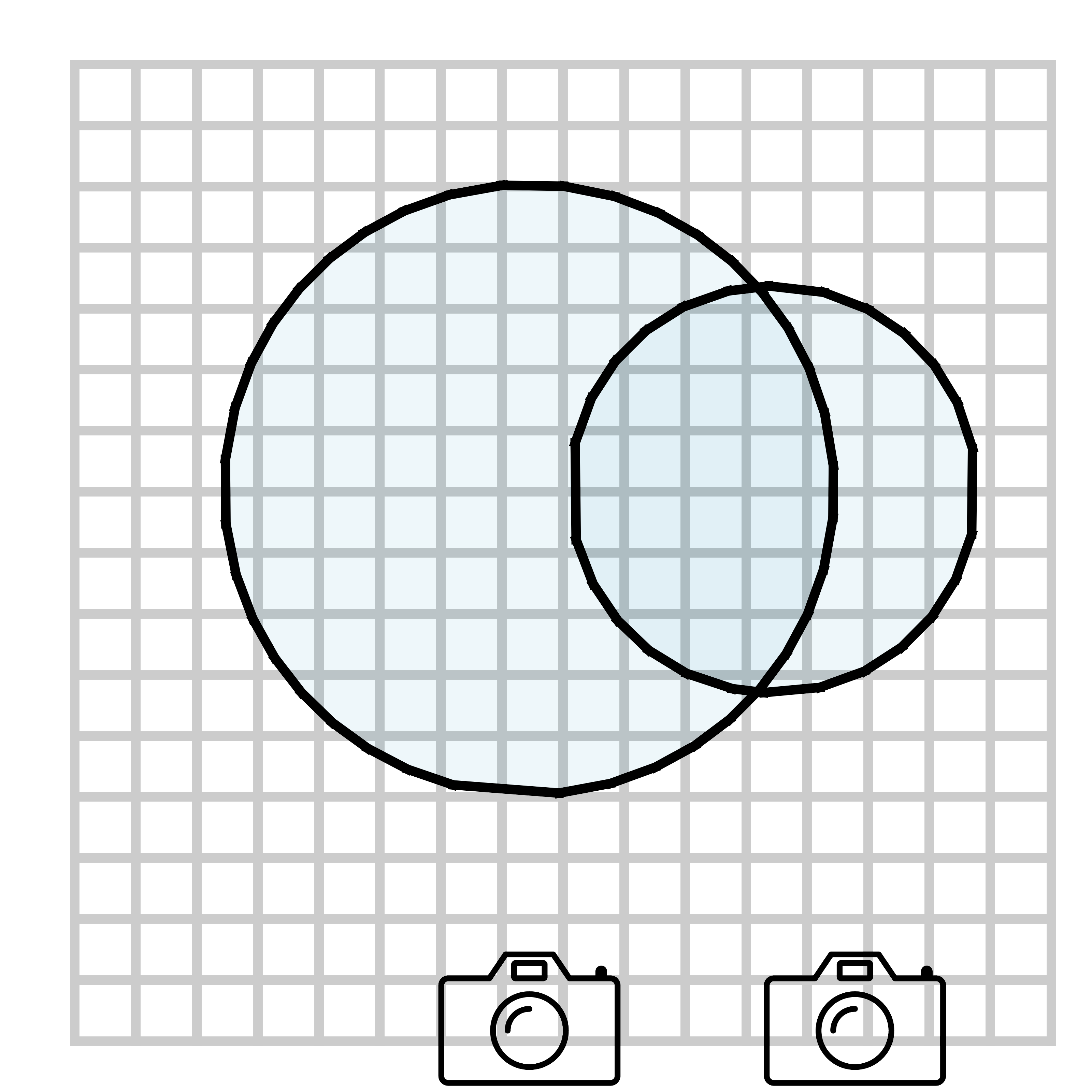}
    \caption*{(a) Uniform grid}
    \end{subfigure}
\begin{subfigure}[t]{.48\linewidth}\centering
    \includegraphics[width=.67\linewidth]{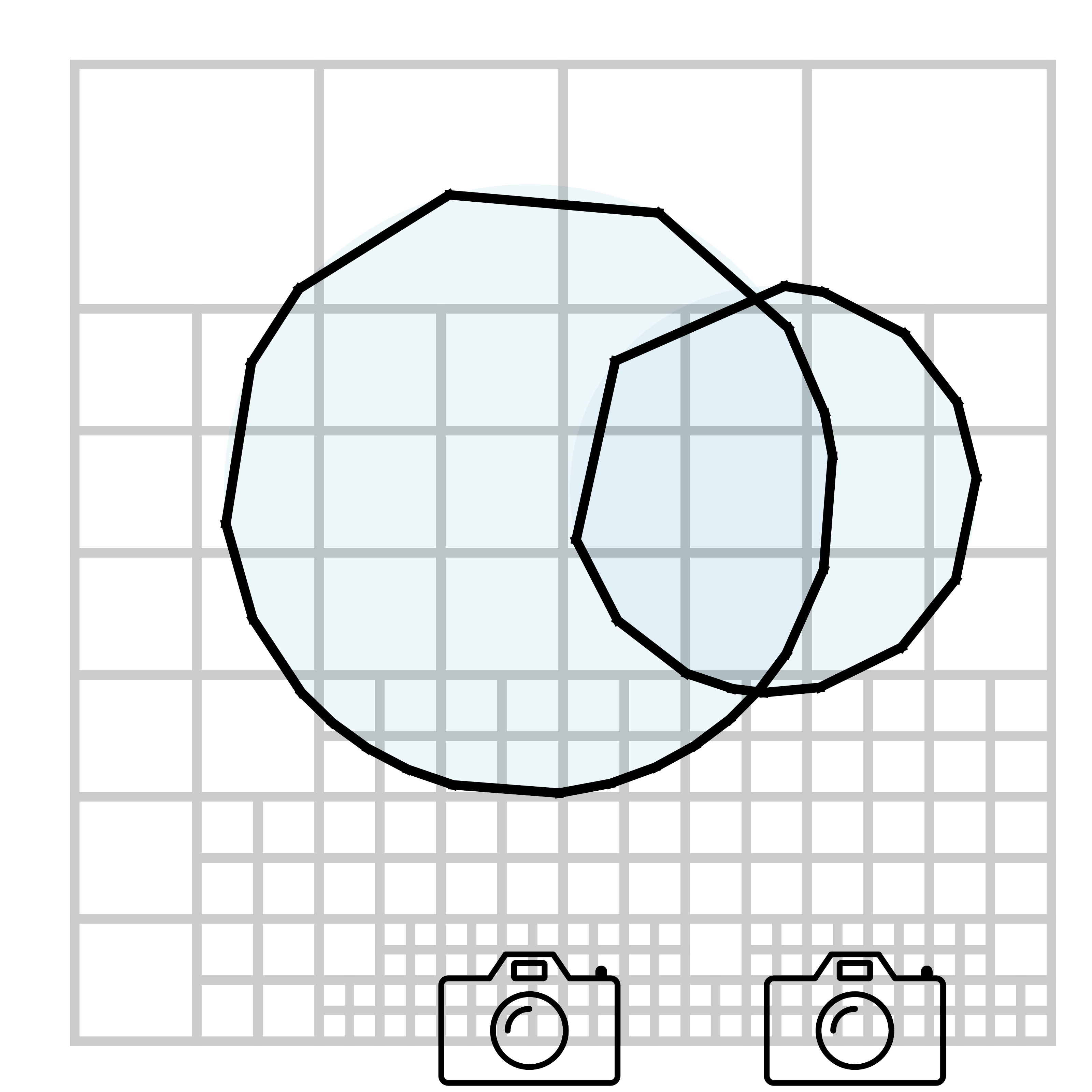}
    \caption*{(b) + Angular diameter}
    \end{subfigure}
\begin{subfigure}[t]{.48\linewidth}\centering
    \includegraphics[width=.67\linewidth]{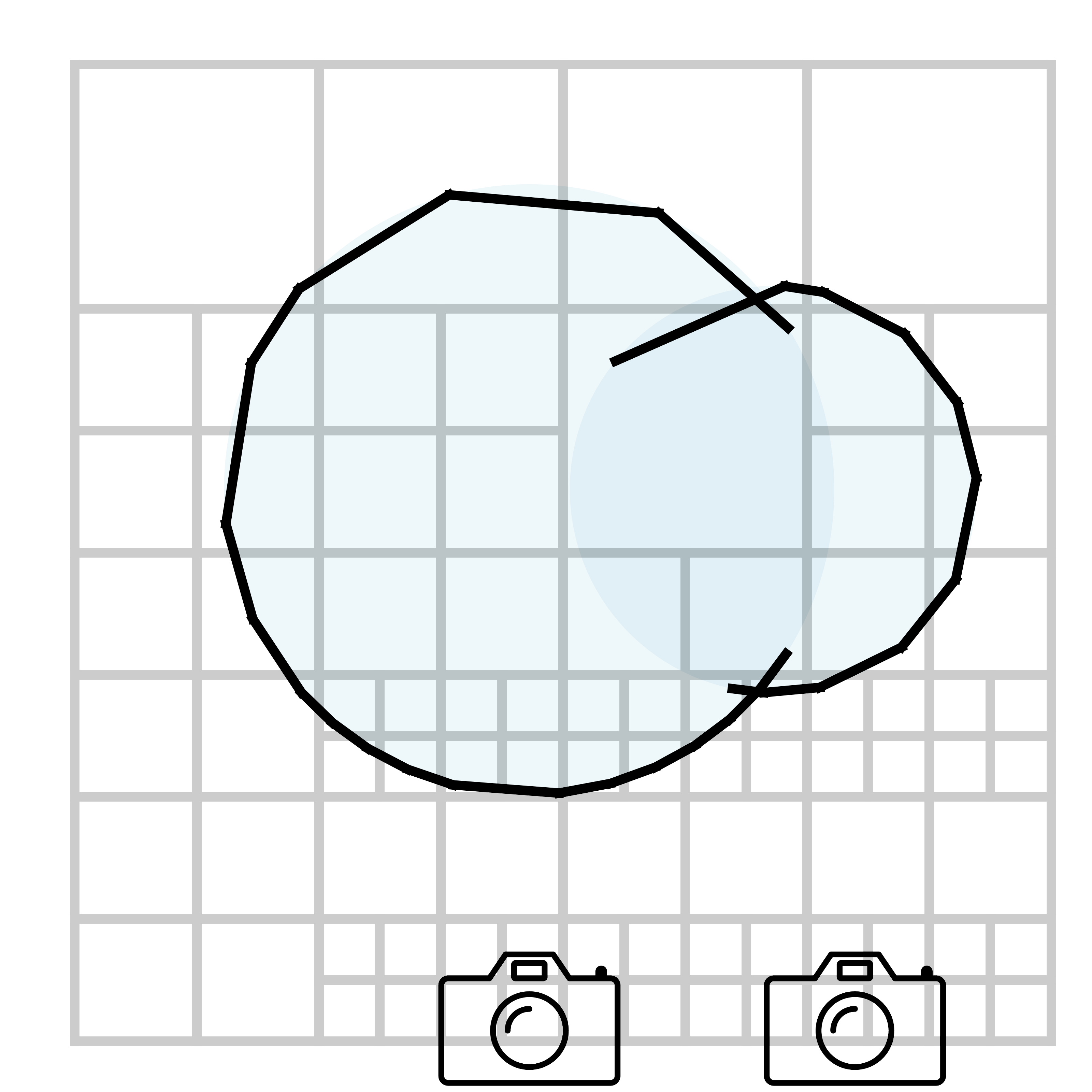}
    \caption*{(c) + Occupancy}
    \end{subfigure}
\begin{subfigure}[t]{.48\linewidth}\centering
    \includegraphics[width=.67\linewidth]{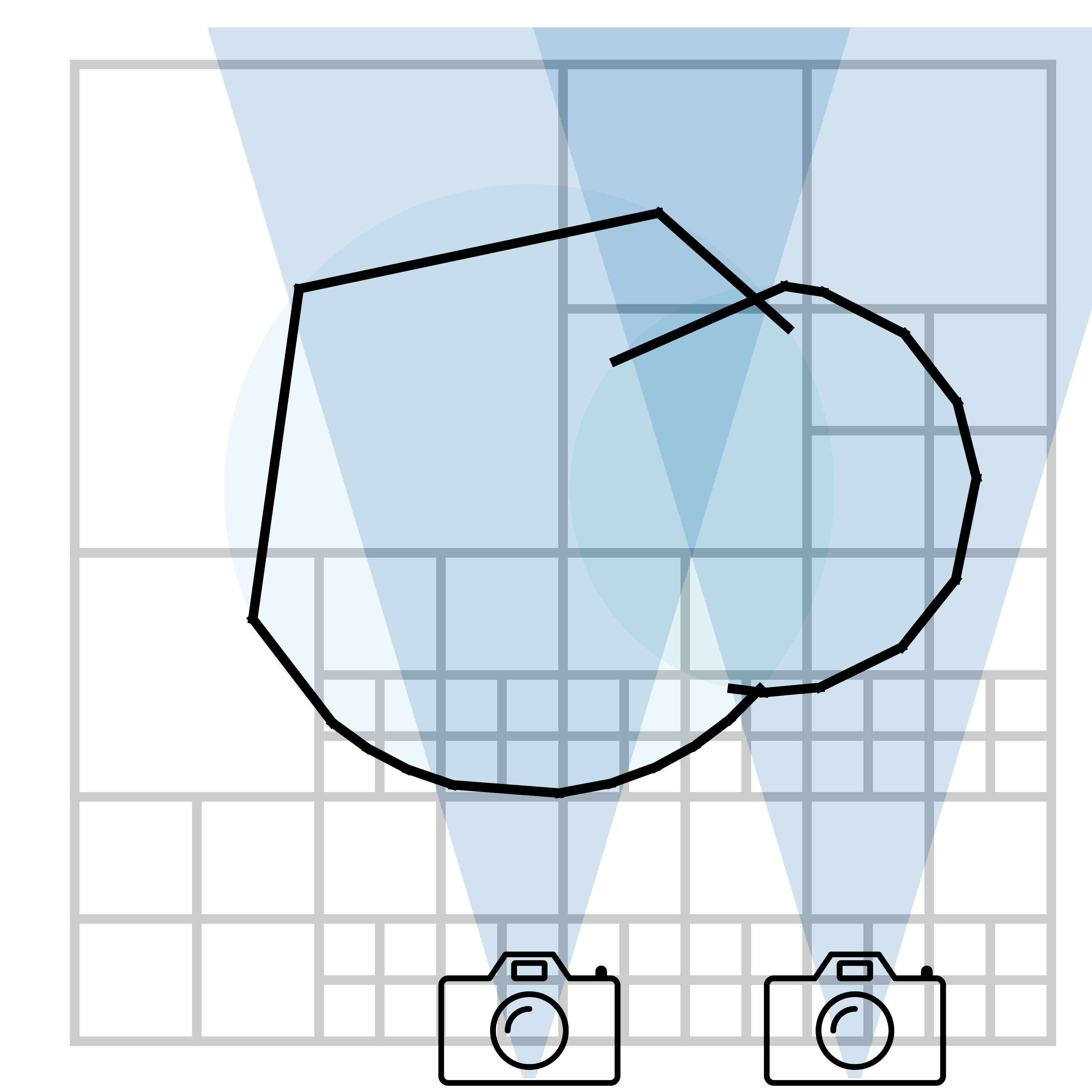}
    \caption*{(d) + Visibility}
    \end{subfigure}
   \caption{The mesh representation (2D analogy) gets increasingly efficient as we apply the 3 LOD criteria one by one.}
   \label{fig:cri}

\end{figure}

\section{Method}

\begin{figure}[b]
  \centering

  \begin{subfigure}[t]{\linewidth}\centering
    \includegraphics[width=0.3\linewidth]{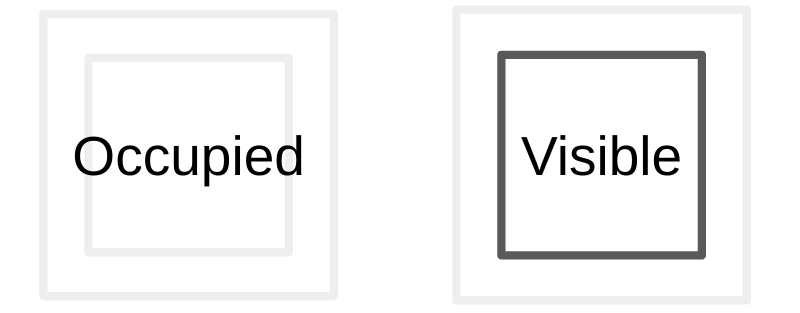}
    \end{subfigure}

\begin{subfigure}[t]{.32\linewidth}\centering
    \includegraphics[width=\linewidth]{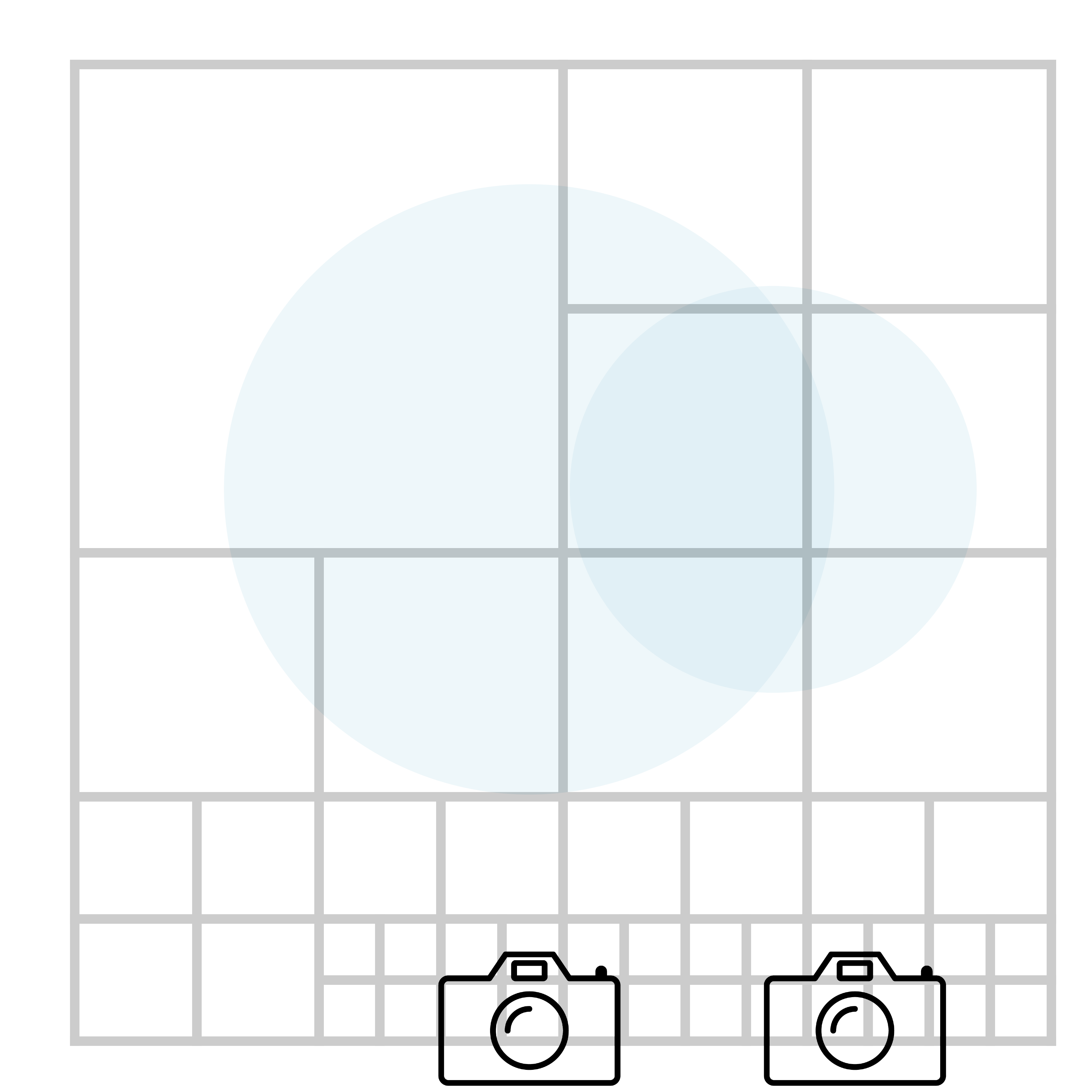}
    \caption*{(a) Coarse full octree}
    \end{subfigure}
\begin{subfigure}[t]{.32\linewidth}\centering
    \includegraphics[width=\linewidth]{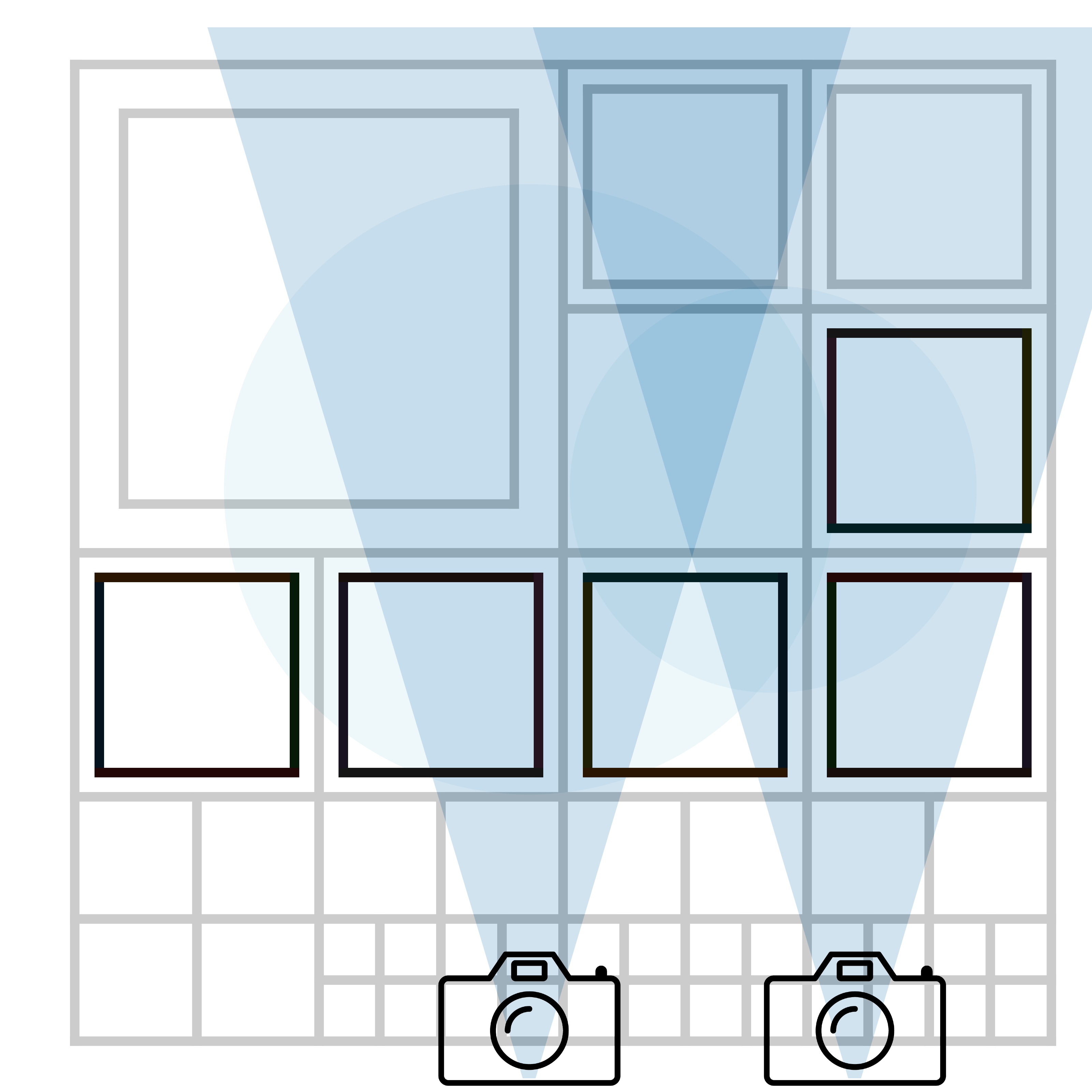}
    \caption*{(b) Occup. and vis. test}
    \end{subfigure}
\begin{subfigure}[t]{.32\linewidth}\centering
    \includegraphics[width=\linewidth]{figures/lodcri3.png}
    \caption*{(c) Fine visible octree}
    \end{subfigure}
   \caption{We construct the octree from coarse to fine.}
   \label{fig:cf}
\end{figure}

Given an SDF and a set of camera poses, we aim to extract an iso-surface mesh that is high-detail when viewed from any camera but has minimum overall polygon-count and rendering cost. Our strategy is to construct an octree by recursively subdividing nodes until several level of detail criteria are met (Sec. \ref{sec:criteria}), using a coarse-to-fine strategy to ensure computing the criterion is feasible (Sec. \ref{sec:coarsetofine}). We provide an overview in Fig. ~\ref{fig:cf}.

\subsection{Octree Level of Detail Criteria}
\label{sec:criteria}

We use an octree to achieve different levels of detail for different parts of the scene. Octrees represent an irregular, multi-resolution grid using a tree structure, where each node represents a cube in the space, and nodes can be recursively subdivided into eight octants. To construct an octree, we recursively subdivide based on the 3 criteria described below and in Fig.~\ref{fig:cri}, using a coarse to fine strategy as described in Sec.~\ref{sec:coarsetofine}.

\vspace{-2mm}

\paragraph{LOD based on Projected Angular Diameter}
The most important LOD criterion is the projected angular diameter of the octree node in the camera views, which is inversely proportional to the distance to the cameras. We compute the maximum angular diameter $A_{\text{node}}$ considering all cameras:

\begin{equation}
    A_{\text{node}} = L_{\text{node}} / \min\limits_{\forall \text{cam}} \text{dist}(\text{node}, \text{cam})\label{eq:angular}
\end{equation}

Where  $L_{\text{node}}$ is the side length of the cube represented by the node, \texttt{cam} iterates through all the cameras, and \texttt{dist} computes the distance of the center of the node to each camera. We compare $A_{\text{node}}$ to our target angular diameter $\hat A$, which is computed  based on the field of view (FOV) of the cameras and the image resolution.
If $A_{\text{node}} > \hat A$, we either subdivide the node into 8 children and compute the angular diameter of each child recursively, or subdivide the node into a denser grid. In practice, there will always be some node containing one of the cameras, and  $A_{\text{node}}$ from Eq.~\ref{eq:angular} can never be less than $\hat{A}$. To avoid this issue, we clamp all distances to be at least $D_{\text{min}}$ (a hyperparameter set by the user), and assert that no is actually within distance $D_{\text{min}}$ of the surface.

\vspace{-2mm}

\paragraph{LOD based on Occupancy}

To save computation, we avoid subdividing any node determined to be wholly unoccupied, as these nodes should not contain zero crossings or affect the final mesh.

Exactly determining whether a node is occupied is expensive. For example, we could completely subdivide every node to meet the angular diameter criterion, and then merge empty nodes, but this is prohibitively expensive. Instead, we determine node occupancy using heuristics based on the SDF value of its 8 bounding vertices, without performing a full subdivision. This misses some nodes that are actually occupied if we further subdivide, and unless handled carefully, will cause dual contouring to create spiky meshes near any sharp LOD transitions. Therefore, we need to propagate the occupancy from the existing surface to its neighbors, as discussed later in Sec. \ref{sec:coarsetofine}.

\vspace{-2mm}

\paragraph{LOD based on Visibility}
Finally, we avoid subdividing nodes which are either occluded by a surface or outside the camera frustrum. In either case, we say that the octree node is not \textit{visible}, since it can only affect the image through shadows, reflections, or bounce lighting. For these not-directly-visible areas, we stop subdivision earlier, i.e. we use $\hat{A}_{\text{inv}} > \hat A$ as the diameter criterion. This reduces overall computation and final mesh size, since the sum of angular sizes of nodes visible to the camera is bounded and should not grow with depth. However, similarly to occupancy checking, computing precise visibility would require completely subdividing all nodes, which is too expensive. We instead decide which node is visible in the middle of the subdivision.

Table~\ref{tab:lod} evaluates the effect of the 3 criteria for the scene in Fig.~\ref{fig:examples} in a low-resolution setting, and for the uniform grid baseline, we could only give the theoretical prediction. We can see all the criteria, especially the angular diameter criterion and the visibility criterion, can significantly reduce redundancy in the octree nodes and the resulting mesh.

\begin{table}[]
   \caption{Ablations on LOD Criteria}
   \label{tab:lod}
\resizebox{\linewidth}{!}{  

\begin{tabular}{llll}
\hline
 & \begin{tabular}[c]{@{}l@{}}\# Octree\\ Leaf Nodes\end{tabular} & \begin{tabular}[c]{@{}l@{}}\# Mesh\\ Vertices\end{tabular} & \begin{tabular}[c]{@{}l@{}}\# Mesh\\ Faces\end{tabular} \\ \hline
 Ang. \xmark \hspace{1mm} Occup. \xmark \hspace{1mm} Vis. \xmark &    $  3.5\times 10^{13} $   &      $ > 1\times 10^{9}        $                                          &                            $> 1\times 10^{9}    $                        \\
 Ang. \cmark \hspace{1mm} Occup. \xmark \hspace{1mm} Vis. \xmark &      8,686,602         &                143,726  &          287,560      \\
 Ang. \cmark \hspace{1mm} Occup. \cmark \hspace{1mm} Vis. \xmark &        1,035,232            &              121,802    &        243,616               \\
 Ang. \cmark \hspace{1mm} Occup. \cmark \hspace{1mm} Vis. \cmark &          736         &          1,535        &     3,079                                                    \\ \hline
\end{tabular}

}
\end{table}

\subsection{Coarse-to-Fine Octree Construction}
\label{sec:coarsetofine}

We use a coarse-to-fine strategy to compute occupancy and visibility criteria efficiently. As illustrated in Fig.~\ref{fig:cf}, we construct the octree in 3 steps: the coarse full octree, the occupancy and visibility test, and the fine visible octree. We explain them step by step.

\paragraph{Coarse Full Octree}
First, we construct a coarse full octree based solely on the projected angular diameter criterion. We initialize the octree with a single root node with size $L_\mathrm{root}$. We use $L_\mathrm{root}=1000\mathrm{m}$ for our Infinigen experiments, but this value is unbounded - one could equally use 5000km (horizon at sea-level) or 50,000km (approx maximum horizon distance on earth) with minimal overhead. We initialize a max-priority queue with the root node, which is sorted by key $A_{\text{node}}$, then repeatedly subdivide the top node while $A_{\text{node}} > \hat{A}_{\text{inv}}$. 

We limit the size of the coarse octree to $S_\text{max}=500,000$. Once this value is reached, we no longer subdivide and instead mark unprocessed nodes as having a dense grid of $N \times N \times N$ ``virtual" children. $N$ is the smallest power of two such that each virtual child has $A_{\text{node}} \le \hat{A}_{\text{inv}}$. Crucially, we avoid storing virtual child nodes in memory, since we can cheaply compute them on the fly given just $A_{\text{node}}$. 

\vspace{-2mm}
\paragraph{Occupancy and Visibility Test}
We use a flood-fill algorithm to avoid querying the occupancy of nodes that are unlikely to be occupied. We start with a subset $\mathcal{N}$ of nodes in the coarse octree (specifically, only the nodes on the boundaries of the $N\times N \times N$ virtual grids). For each node in $\mathcal{N}$, we compute the SDF value on the 8 vertices of the node. If both positive ($\ge 0$) and negative ($< 0$) vertices exist, we mark this node as occupied and put its neighboring nodes into $\mathcal{N}$. We iterate until we completely test $\mathcal{N}$. Because the iso-surface of the SDF is continuous, we will not miss a node if any node in its connected component is checked. This approach may miss small floating islands (i.e. if they dont intersect the corner of any coarse node), but we accept this as a tradeoff.

Next, we compute whether each occupied node is also visible or not. For each camera, we project all nodes onto a depth buffer and mark nodes as visible in this camera if the projected depth is smaller than any of the values within a neighborhood of its projected pixel. A node is then visible overall if it is visible in at least one camera camera. To further avoid missing potentially visible nodes in the final result, we dilate the set of visible nodes by including all $<k$-th degree neighbors, where $k=2$ is a user-specified hyperparameter.

\vspace{-3mm}
\paragraph{Fine Visible Octree}

Finally, we divide all visible nodes into high resolution octrees until all of these octrees's leaf nodes (referred to as subnodes) have $A_{\text{subnode}} \le \hat{A}$. This process can cause sharp LOD transitions if a neighboring node that was not considered occupied turns out to be occupied. Therefore, if we find a sign change while dividing a node on the border of a node marked unoccupied, we mark it as occupied such that it is meshed at high resolution too. This propagation without restriction can undo the work done by the visibility test. Therefore if a neighbor node is already marked as invisible, we don't convert it to be visible.

\subsection{Mesh Extraction}
We use the dual contouring algorithm~\cite{perry2001kizamu,ju2002dual,ju2006intersection} to extract a mesh for each component from the octree with the $\text{SDF}_{\text{comp.}}$ on each vertex. But, instead of quadratic error functions (QEF), we use bisection to locate the center vertex of each node, which is more robust to discontinuous SDFs.

\subsection{Computational Complexity}
The majority of the time and the memory are spent in the fine octree step, where we need to densely query the SDFs and extract dense meshes. Such SDFs are usually optimized for parallel computation. Therefore we do these operations in batches on parallel devices such as GPU to save time. The total complexity  depends on many factors:
\vspace{-1mm}

\begin{equation}
\text{Time and Memory} \propto K_{\text{scene}} K_{\text{cam}} N_{\text{cam}} \frac{S_{\text{fov}}}{\hat{A}^2}
\end{equation}

\vspace{-1mm}

Where $K_{\text{scene}}$ is the complexity of the scene, $K_{\text{cam}}$ describes how fast the camera moves, i.e., how much new content each camera has (for completely non-overlapping cameras, $K_{\text{cam}}=1$, but in practice, $K_{\text{cam}} \ll 1$), $N_{\text{cam}}$ is the number of cameras, and $S_{\text{fov}}$ is the solid angle of the FOV. The last factor $\frac{S_{\text{fov}}}{\hat{A}^2}$ comes from the fact that we only construct the fine octree for the visible part.

\begin{figure*}[t]
  \centering
  \begin{minipage}{0.48\textwidth}
    \includegraphics[width=\linewidth]{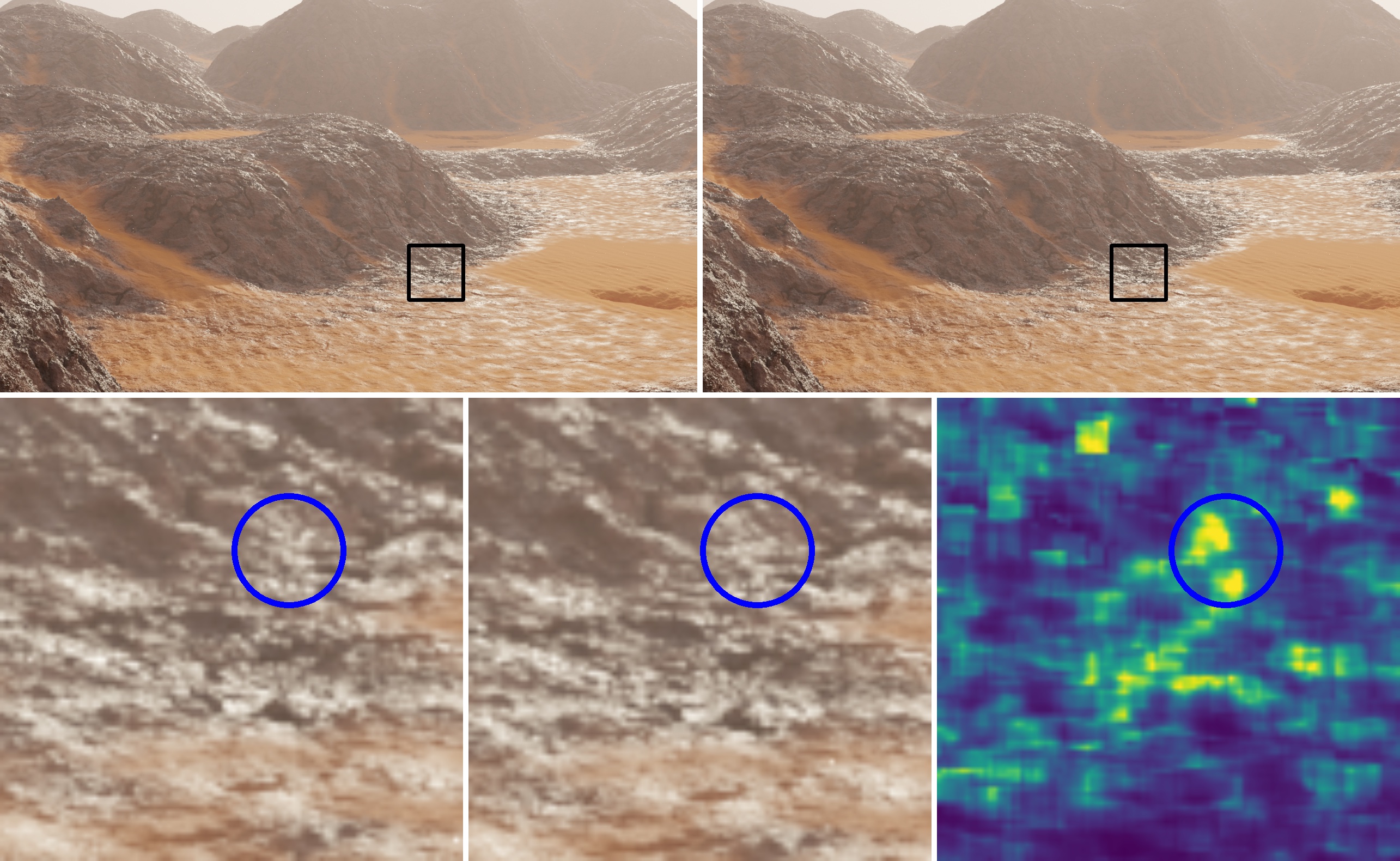}
    \caption*{(a) \infinigen{}}
    \end{minipage}
  \begin{minipage}{0.48\textwidth}
    \includegraphics[width=\linewidth]{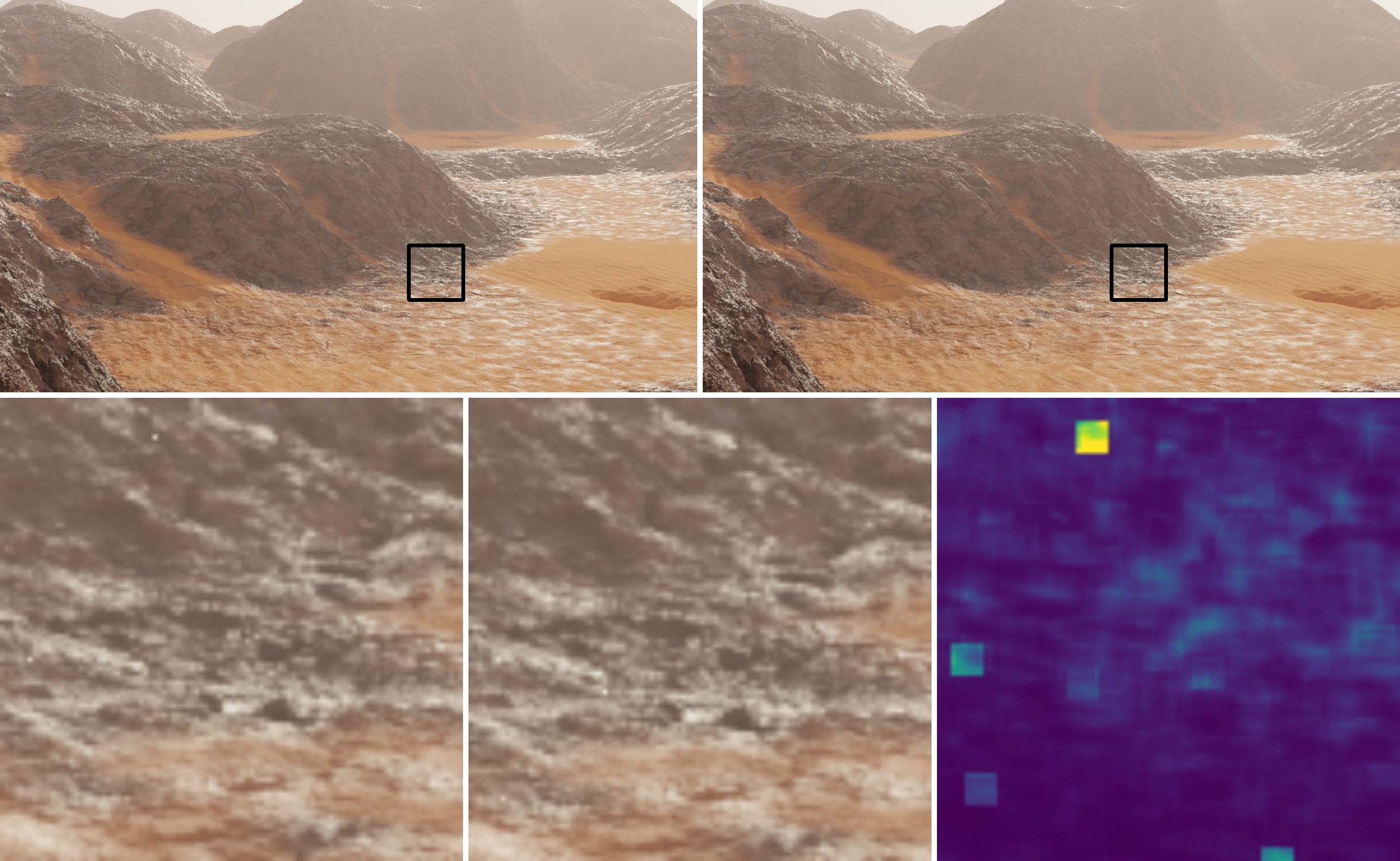}
    \caption*{(b) Ours}
    \end{minipage}

   \caption{ Two adjacent frames of meshes extracted with (a) The Infinigen solution (b) \projectname{} (Ours). In the zoomed-in images, we can see more visual inconsistency in (a). The heatmap shows the quantitative measurement (via flow ground truth, explained in Sec.~\ref{sec:temp}).}
   \label{fig:examples}

\vspace{-2mm}
   
\end{figure*}

\begin{figure*}[!t]
  \centering
      \begin{subfigure}[t]{.19\linewidth}\centering
\includegraphics[width=\linewidth]{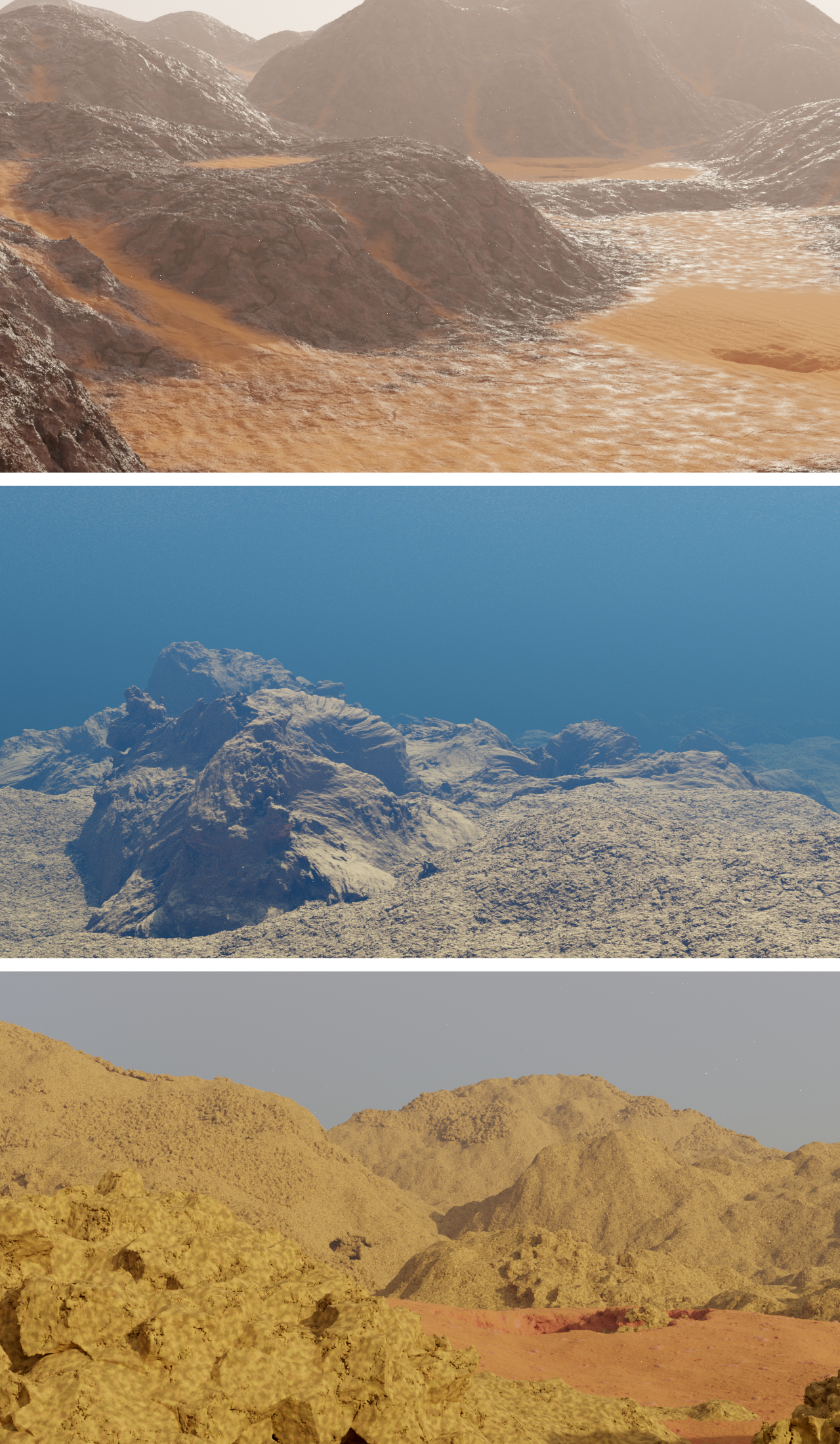}
    \caption*{(a) Scene overview}
    \end{subfigure}
        \begin{subfigure}[t]{.57\linewidth}\centering
\includegraphics[width=\linewidth]{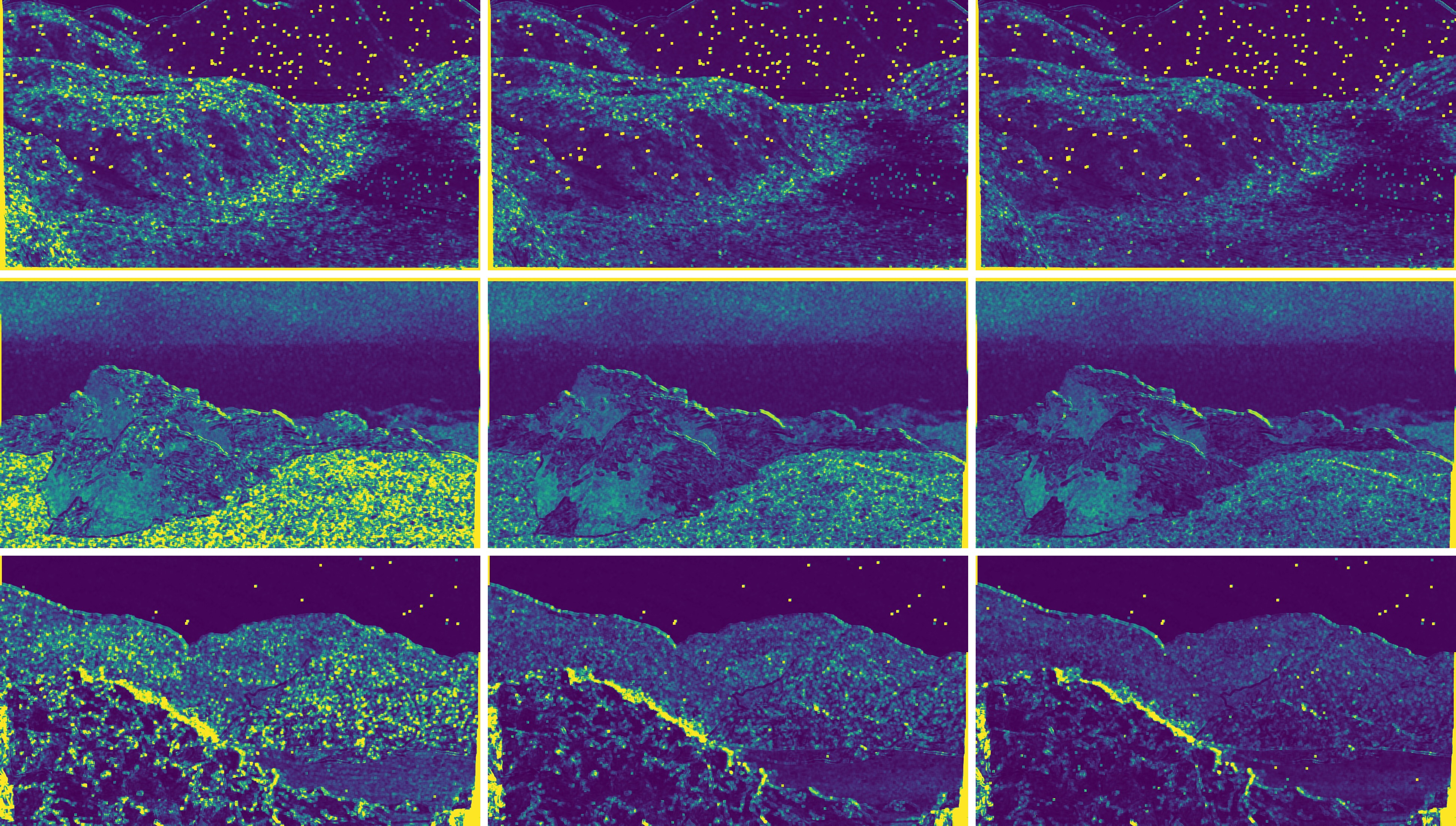}
    \caption*{(b) \infinigen{} with increasing resolution}
    \end{subfigure}
        \begin{subfigure}[t]{.19\linewidth}\centering
\includegraphics[width=\linewidth]{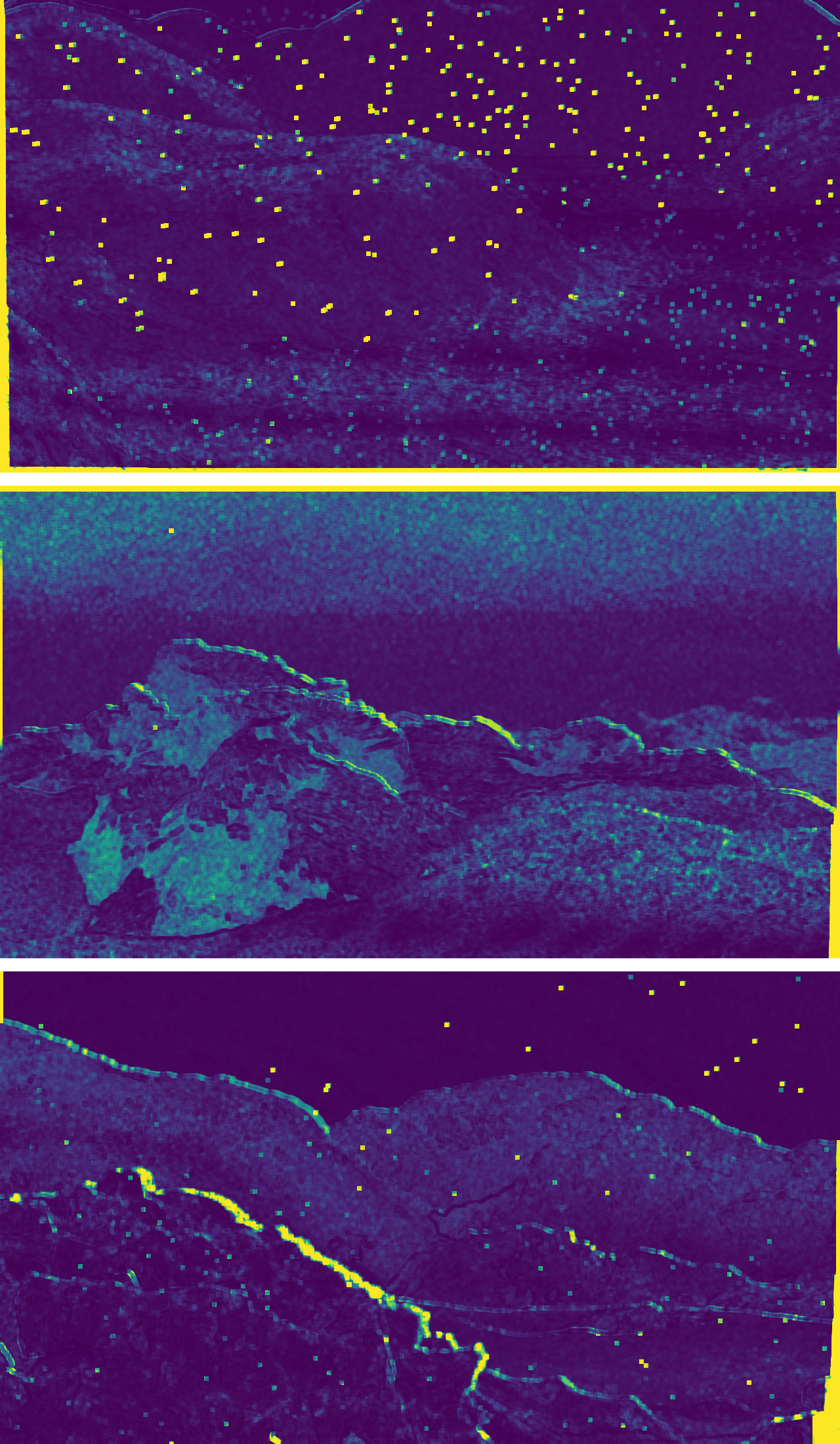}
    \caption*{(c) Ours}
    \end{subfigure}

   \caption{Quantitative measurement of view consistency for the first pair of transition frames. The brighter, the worse.}
   \label{fig:temp}
\end{figure*}

\begin{figure*}[!t]
  \centering
        \begin{subfigure}[t]{.48\linewidth}\centering
\includegraphics[width=\linewidth]{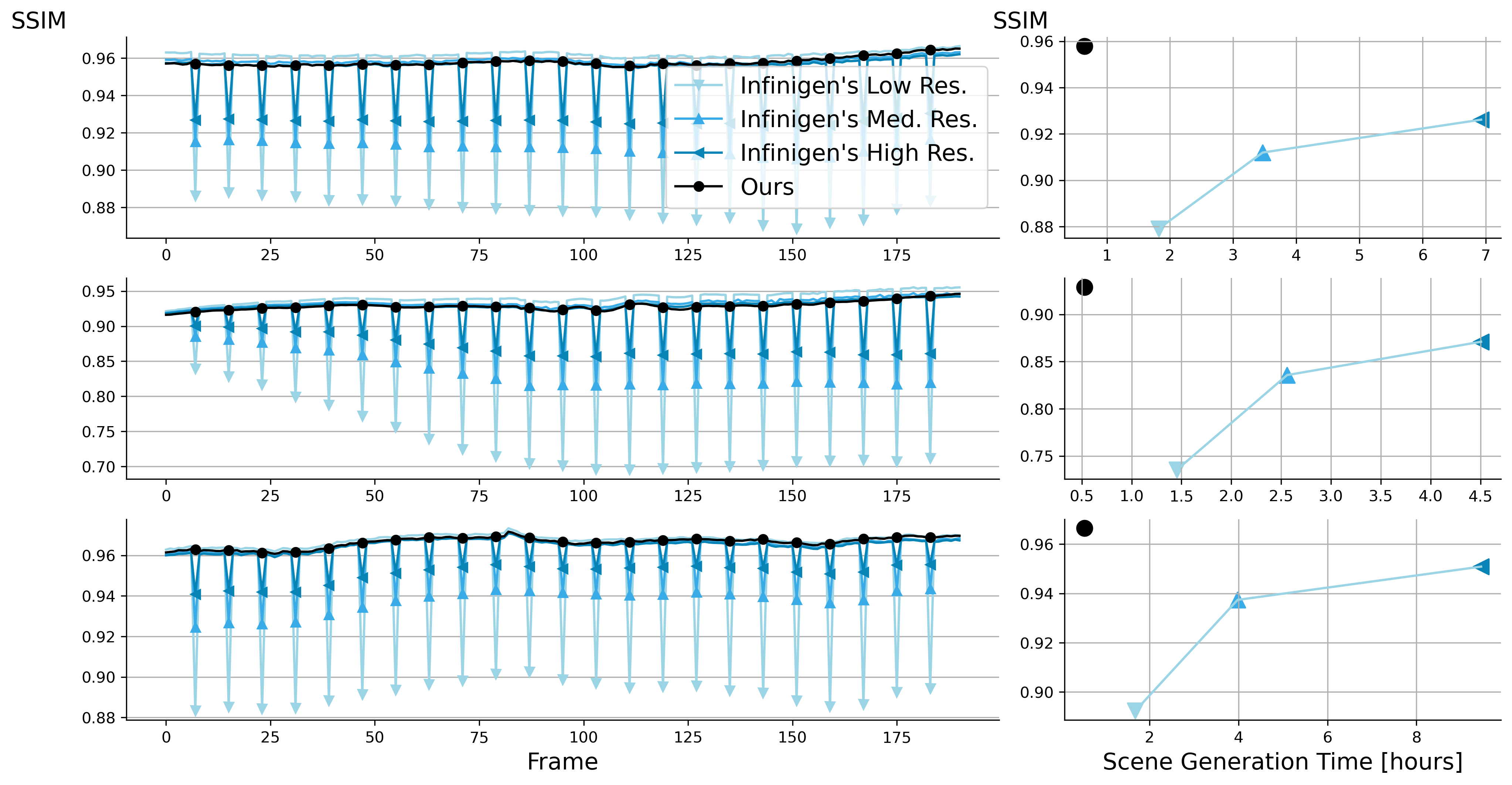}
    \caption*{(a) Comparison of frame-by-frame SSIM}
    \end{subfigure}
        \begin{subfigure}[t]{.48\linewidth}\centering
\includegraphics[width=\linewidth]{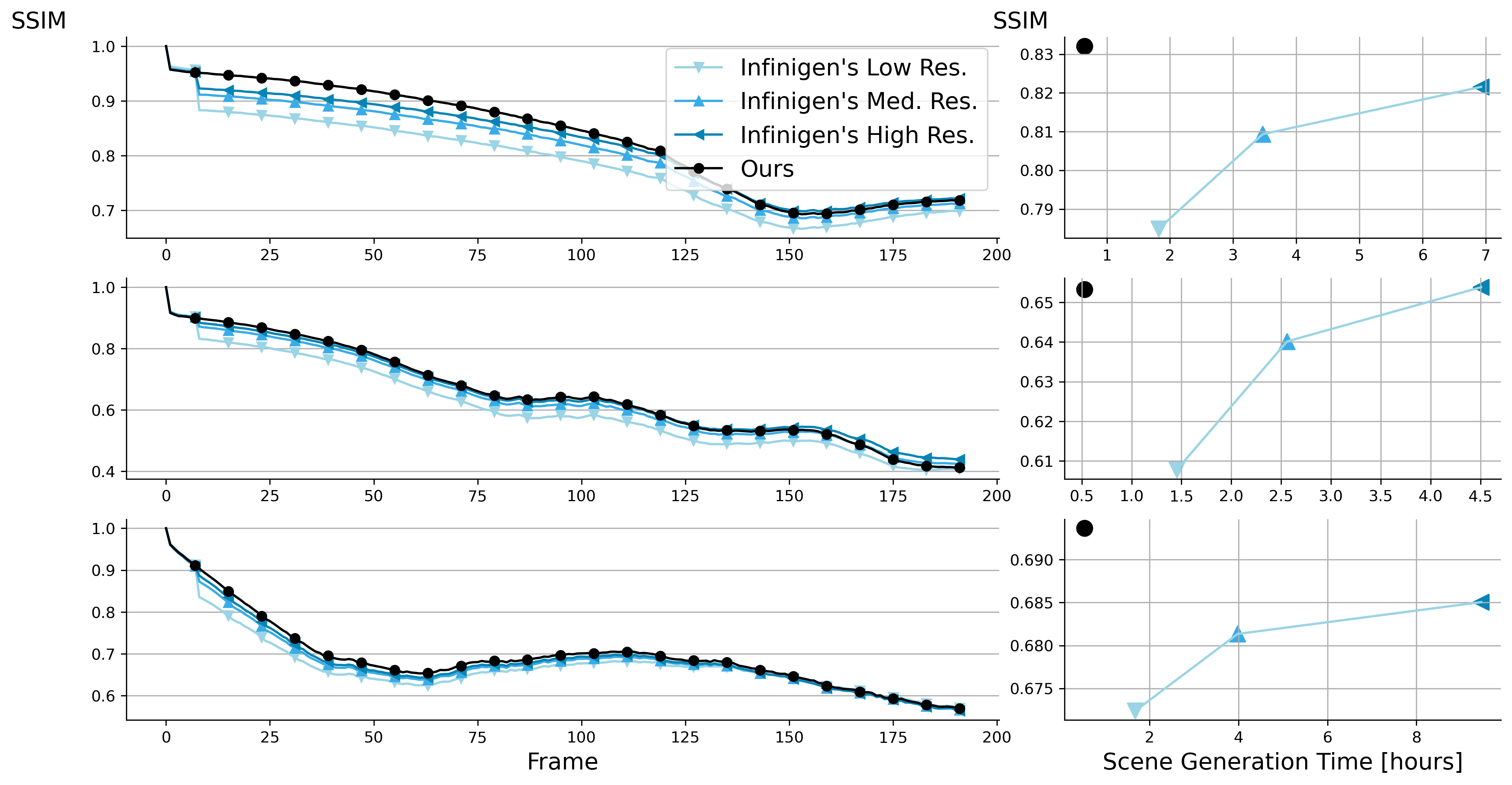}
    \caption*{(b) Comparison of first-to-nth frame SSIM}
    \end{subfigure}

   \caption{Quantitative measurement of view consistency (a) for adjacent frames and (b) along a pixel trajectory. On the left, Infinigen's transition frames introduce sharp spikes in frame-by-frame inconsistency. On the right, Infinigen suffers worse $0 \rightarrow i$ consistency except for the first 8 before any transition frames occur. In both cases, our method provides a superior SSIM-vs.-runtime tradeoff.}
   \label{fig:ssim_tradeoff}
   \vspace{-2mm}
\end{figure*}

\section{Experiments}
To generate synthetic scenes and videos, we integrated our method with Infinigen~\cite{raistrick2023infinite}'s scene generator but replace their mesh extracting solution with ours. Because Infinigen's assets besides terrains do not use SDFs, we turn them off in the experiments.
We randomly generated 42 scenes and for each scene, we rendered a video consisting of $N=192$ frames of resolution $H\times W=720\times1280$ on a cluster of GPUs including NVIDIA GeForce RTX 3090, RTX A6000 and A40 . Each scene is generated under several settings:

\begin{itemize}
    \item With the original Infinigen solution, at 3 target resolutions. Higher mesh resolution reduces flickering, but incurs more computational cost. We show the 3 target resolutions to demonstrate this trade-off. We re-generate the mesh every 8 frames, which is the default setting from Infinigen and is a generous comparison due to the low-poly artifacts at image edges during camera motion, as shown in Fig.~\ref{fig:scene_contrast}.
    \item With our method using a moderate resolution. We create a single mesh for all the $N$ frames except for when the resulting mesh exceeds Blender's (our rendering Engine) capacity, which happens for 5 scenes out of the 42 scenes in total. In those cases, we split the entire video clip into two clips. This makes the video of these 5 scenes flicker in the middle frame, but it has negligible effects on the experiment results compared with the theoretical results.
\end{itemize}

We render RGB images with Blender's Cycles Engine, with 8192 samples per pixel. We also save ground-truth depth maps and camera parameters for each frame. Because we only consider static terrain, we can compute the ground truth optical flow $\mathbf{F}_{i\rightarrow j}$ ($1 \le i,j \le N $) for any pair of frames \# $i$ and \# $j$.
Each vector $\mathbf{F}_{i\rightarrow j}[x,y]$ means pixel $(x, y)$ in frame \# $i$ goes to pixel $(x, y) + \mathbf{F}_{i\rightarrow j}[x,y]$ in frame \# $j$.

\subsection{View Consistency}
\label{sec:temp}

First, we show that our method has less flickering in the rendered videos, i.e., ours has better view consistency. Qualitatively, we show a zoomed-in region of a muddy mountain scene for a pair of mesh-switching frames (we call such frames transition frames) rendered with both methods in Fig.~\ref{fig:examples}. Because the \infinigen{} solution switches between meshes, you can see the differences between the two images, focusing on those reflective regions.

In addition, we can quantitatively measure the view consistency score $S_{i\rightarrow j}$ between two frames $i$ and $j$ given the ground truth optical flow $\mathbf{F}_{i\rightarrow j}$:

\begin{equation}
S_{i\rightarrow j} = \text{SSIM}(\mathbf{I}_{i}, \text{warp}(\mathbf{I}_{j}, \mathbf{F}_{i\rightarrow j})) \label{eq:ssim}
\end{equation}
\vspace{-2mm}

Where $\mathbf{I}_{i}$ and $\mathbf{I}_{j}$ are the RGB images of frame $i$ and frame $j$ respectively, \text{warp} is a function that warps back the input frame according to the forward flow, and \text{SSIM} computes the structural similarity score (SSIM) ~\cite{wang2004image} of two images. This produces a 2D map the same size as the image, with values proportional to how consistent the two frames are. Fig.~\ref{fig:examples} shows the SSIM map for the zoomed-in regions. 

We analyze the complete video sequence for 3 scenes, and provide more in the supplementary material. In Fig.~\ref{fig:temp}, column (a) shows an overview of the scene for reference; column (b)(c) shows the SSIM score map $S_{8\rightarrow 9}$ between the first pair of mesh-switching frames, i.e., frame \#8 and frame \#9. Column (b) shows 3 different resolution settings with the \infinigen{} solution; (c) shows ours. The brighter the color, the worse the SSIM score. From left to right in (b), as the resolution increases, the score gets better. Yet, ours has much better view consistency even compared with the best in \infinigen{}. The only significant inconsistency (yellow area) lies in occlusion boundaries and where volume scattering makes the rendering noisy.

Fig.~\ref{fig:ssim_tradeoff}~(a) shows the SSIM score for adjacent frames, which details how much flickering is experience when we watch the video continuously. It shows the frame-wise average against time on the left. The sharp comb-like peaks indicate periodical flickering in the Infinigen solution during transition frames. On the right, we show scene-wise average end-point-error (EPE) for transition frames versus the mesh extraction time. We do not include rendering time because it is very dependent on the rendering engine. \infinigen{}'s flickering is reduced by additional mesh resolution, but this incurs increased runtime. However, even in its highest resolution, it is still much worse than our method.

Fig.~\ref{fig:ssim_tradeoff}~(b) shows SSIM score along many individual pixel trajectories, comparing frame 1 to frame $i$ at every timestep. It also shows the frame-wise average against time and the scene-wise average against the generation time. We can see as soon as the frame goes beyond the first group of frames where the mesh is reused, the score drops significantly in Infinigen, so it is much harder to match corresponding points there. The trade-off curve is similar.

\begin{figure}[t]
  \centering
  \begin{subfigure}[t]{\linewidth}\centering
    \includegraphics[width=\linewidth]{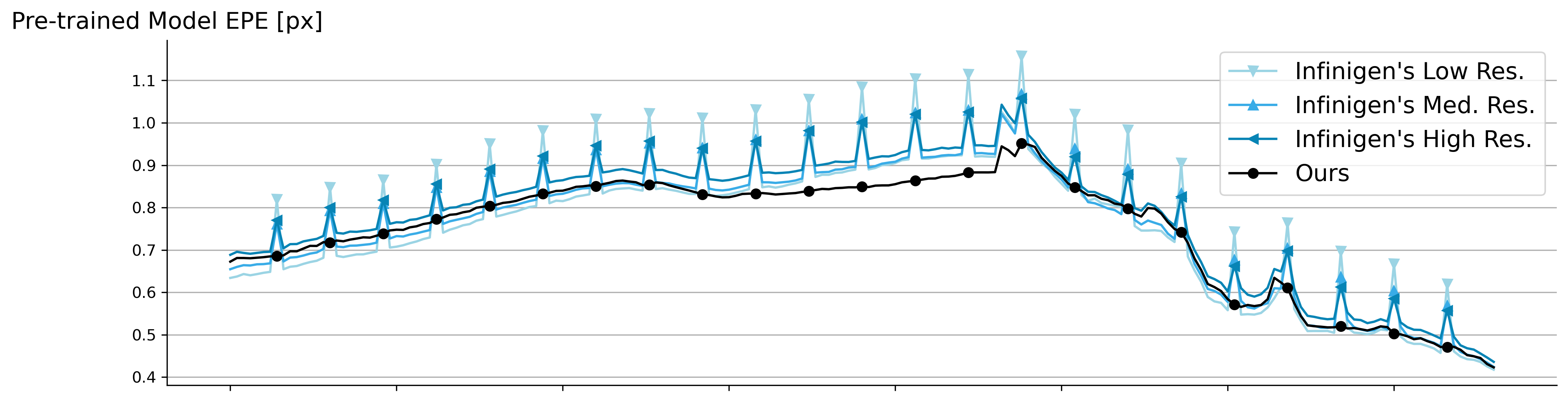}
    \caption*{(a) Gunnar Farneback's algorithm}
    \end{subfigure}
      \begin{subfigure}[t]{\linewidth}\centering
    \includegraphics[width=\linewidth]{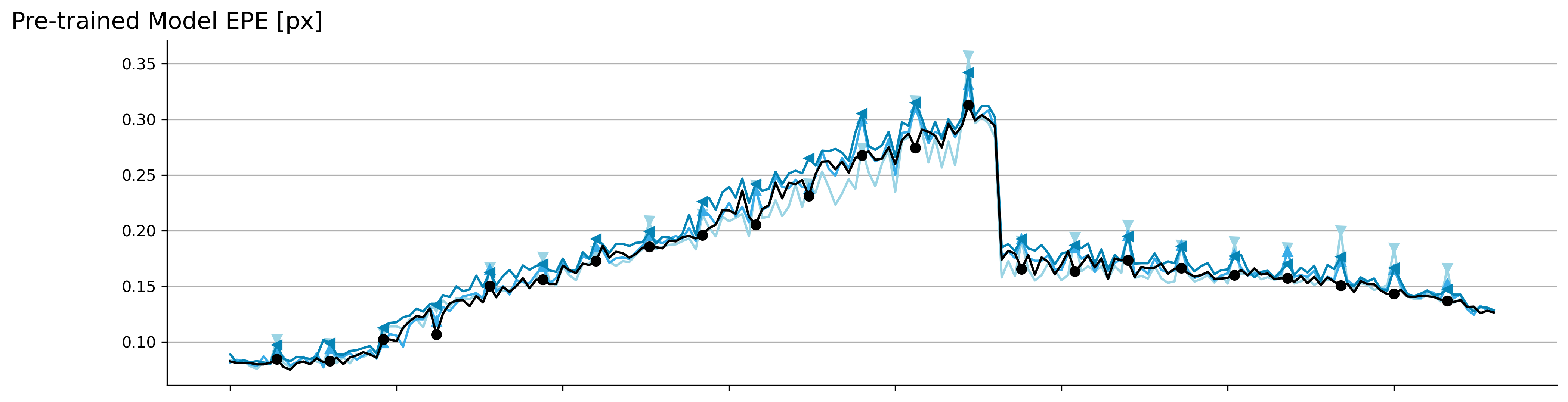}
    \caption*{(b) RAFT}
    \end{subfigure}
          \begin{subfigure}[t]{\linewidth}\centering
    \includegraphics[width=\linewidth]{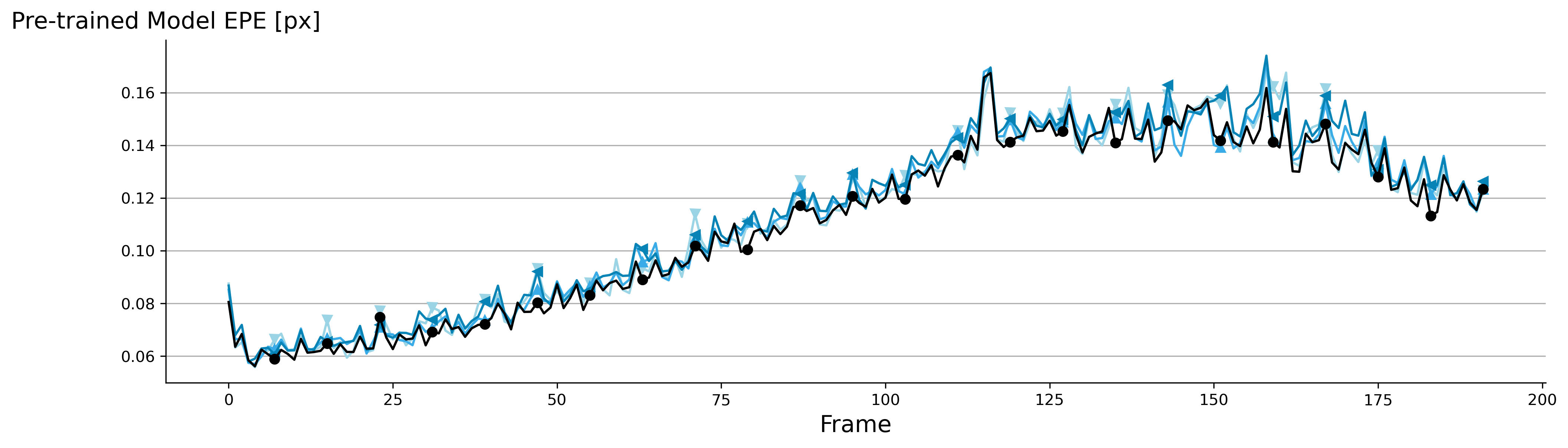}
    \caption*{(c) VideoFlow}
    \end{subfigure}
    
   \caption{End-point-error (EPE) for 3 pre-trained optical flow methods evaluated on meshes from \projectname{} vs. Infinigen at various resolutions.}
   \label{fig:flow_benchmark1}
   \vspace{-2mm}
\end{figure}

\begin{figure}[t]
  \centering
   \vspace{-2mm}
    \includegraphics[width=0.8\linewidth]{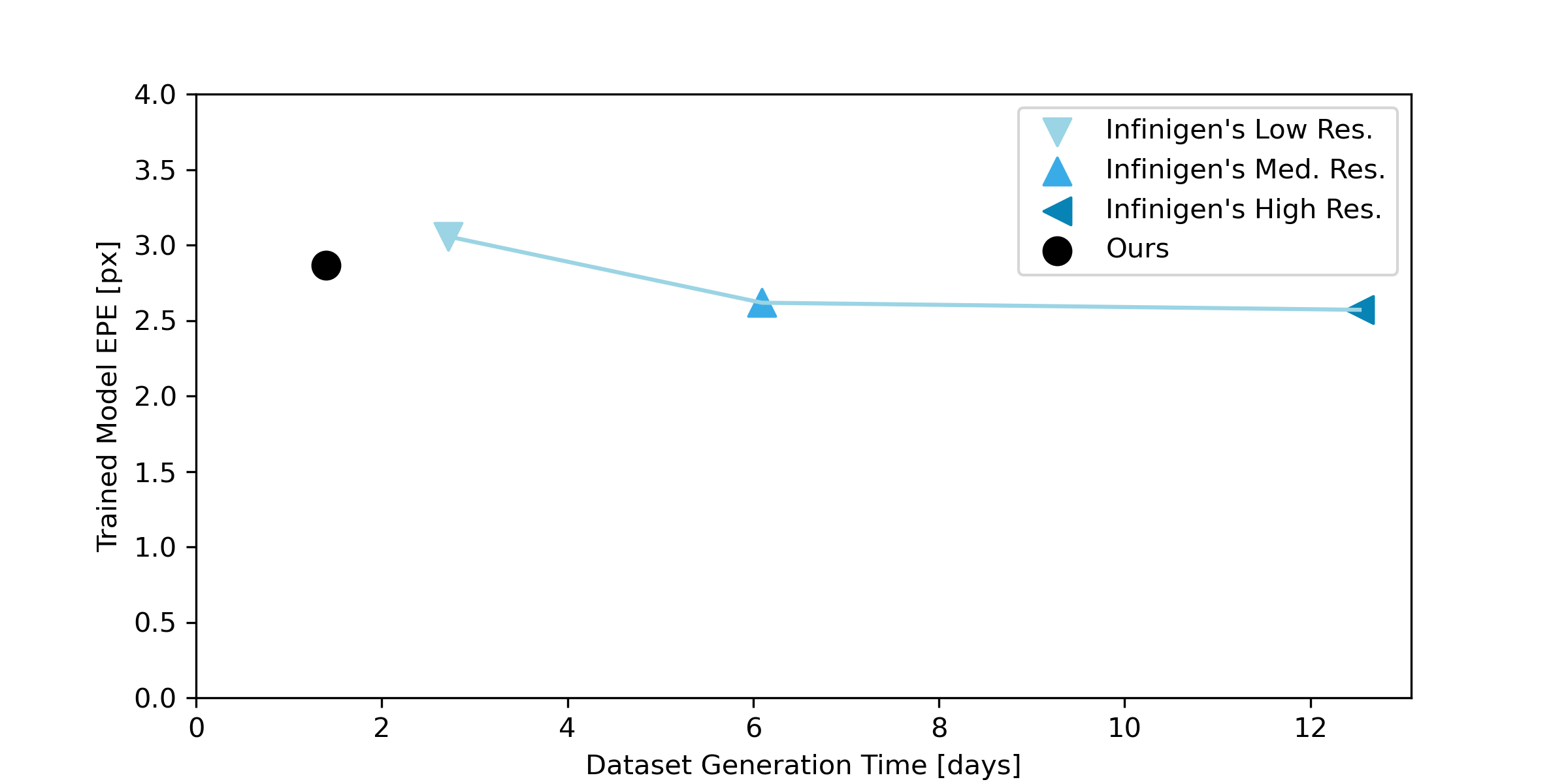}
    \includegraphics[width=0.8\linewidth]{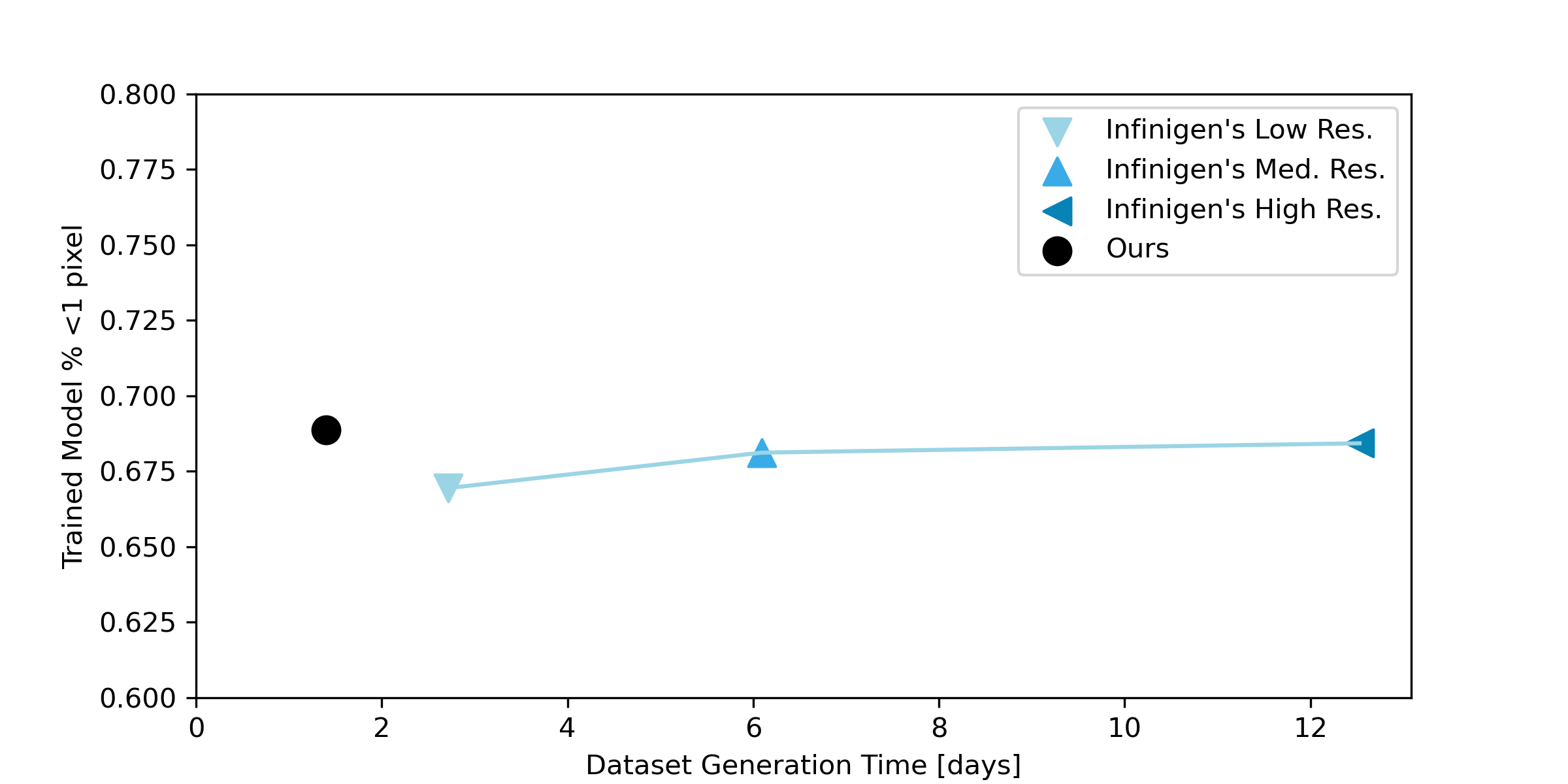}
   \caption{Models trained on our dataset achieve better or comparable results with a lower cost.}
   \label{fig:training}
   \vspace{-3mm}
\end{figure}

\begin{figure*}[t]
  \centering
  \begin{minipage}{0.48\textwidth}
    \includegraphics[width=\linewidth]{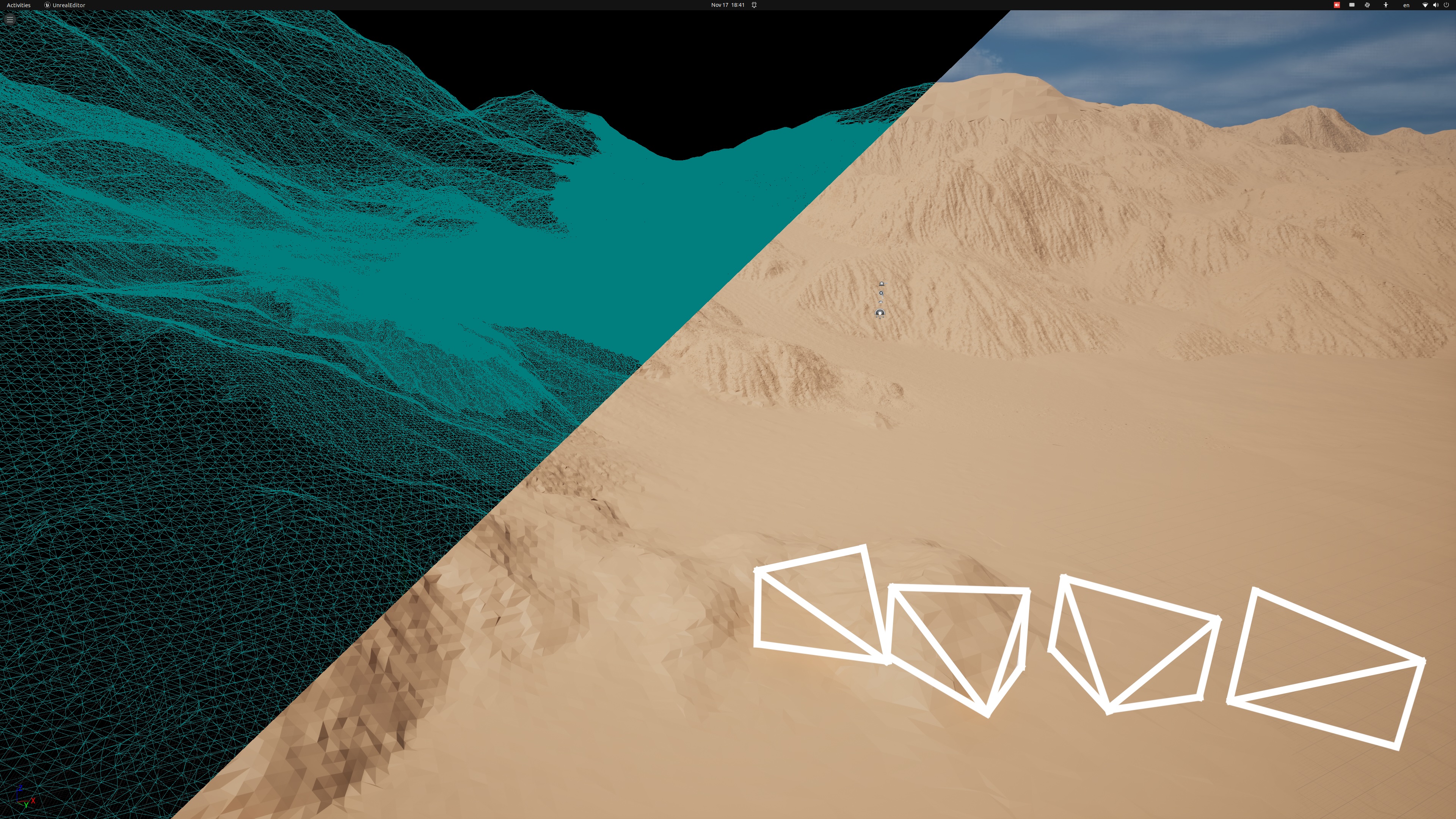}
    \caption*{(a) The scene and wireframe visualization}
    \end{minipage}
  \begin{minipage}{0.48\textwidth}
    \includegraphics[width=\linewidth]{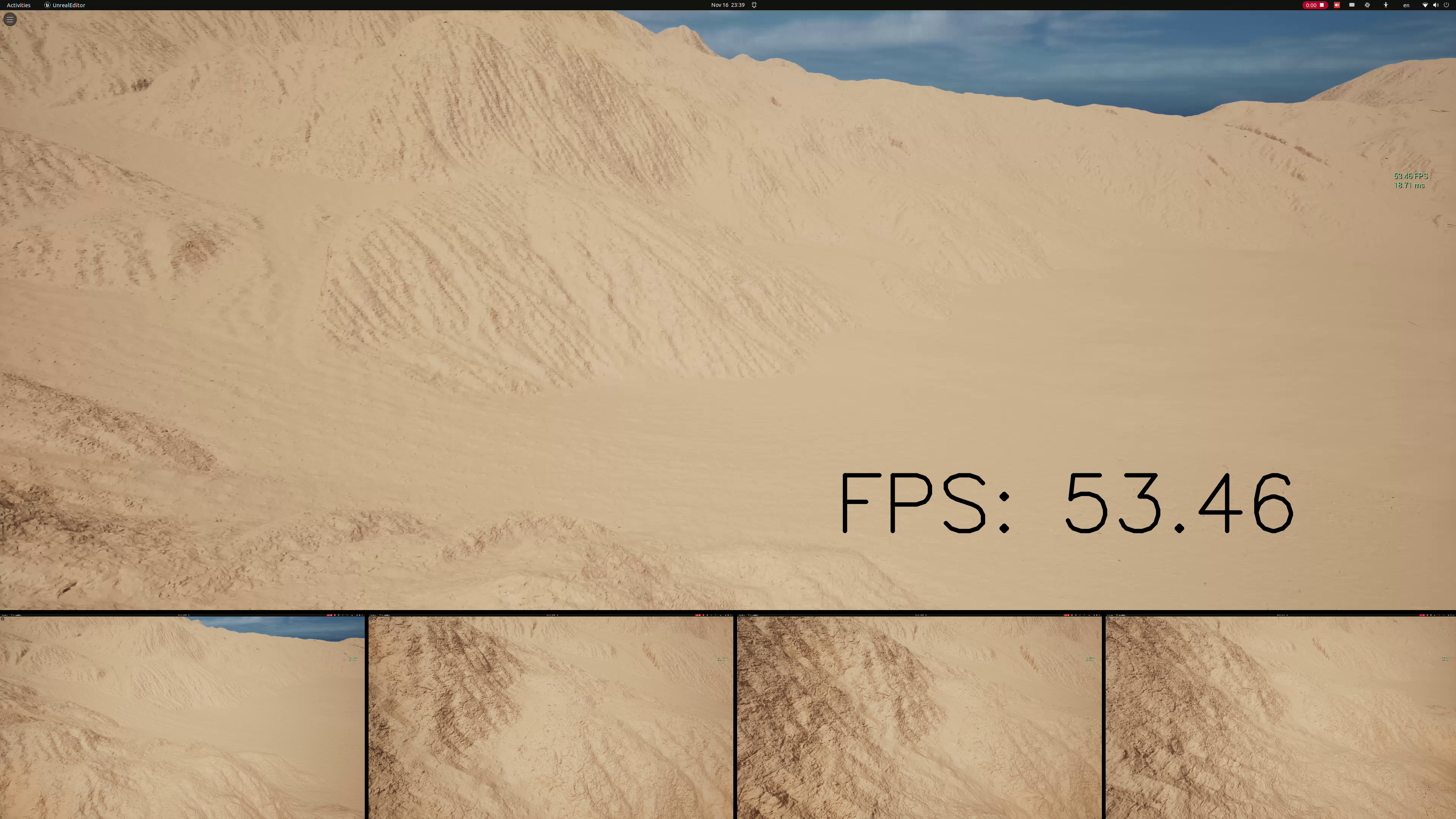}
    \caption*{(b) Screenshots of a real-time video. FPS: 53.46}
    \end{minipage}

   \caption{ Our method enables real-time (\textgreater 50 FPS) scene exploration in Unreal Engine}
   \label{fig:realtime}
   \vspace{-2mm}
\end{figure*}

\subsection{Improved Benchmarks}

We compare these datasets as benchmarks for the optical flow estimation task. With the same method and the same scene, if a dataset setting produces stable and reasonable errors as the camera moves smoothly, this means the dataset has fewer distracting factors and can serve as a better benchmark. We evaluate optical flow for each pair of adjacent frames using 3 existing methods: Gunnar Farneback's algorithm~\cite{farneback2003two} from OpenCV, RAFT~\cite{teed2020raft}, and VideoFlow~\cite{shi2023videoflow}. Fig.~\ref{fig:flow_benchmark1} shows the results of each method for one scene (see the supplement for more). On the left, we show frame-wise mean endpoint error (EPE). Again,we see spikes during transition frames in \infinigen{}, especially in the first two methods which estimate flow independently without neighbor information. This demonstrates that \infinigen{} has significant measurement noise when used as a benchmark dataset. In contrast, evaluating on our data produces consistent and smooth errors, except for the middle frame where the camera motion suddenly changes direction, which is an expected phenomenon.

\subsection{Improved Model Training}

We also use the generated datasets as training sets for the optical flow task. Particularly, we focus on wide baseline scenarios, which is a more challenging task and more useful for relevant applications like stereo matching. We use RAFT~\cite{teed2020raft}, and train the model from scratch for 50k iterations with batch size 2 on image pairs from 32 out of the 42 scenes, with the rest as a validation set. We train on images sampled 10 frames apart, and exclude images with median optical flows greater than 50px.  We compare the training results versus the mesh generation cost of the dataset in Fig~\ref{fig:training}.  We can see that models trained on our dataset attain better EPE and \% \textless 1 px metrics and at a lower lower cost. \infinigen{} is prohibitively expensive to achieve similar performance.

\subsection{Creating Embodied AI Environments}
\begin{figure}[t]
  \centering
    \includegraphics[width=\linewidth]{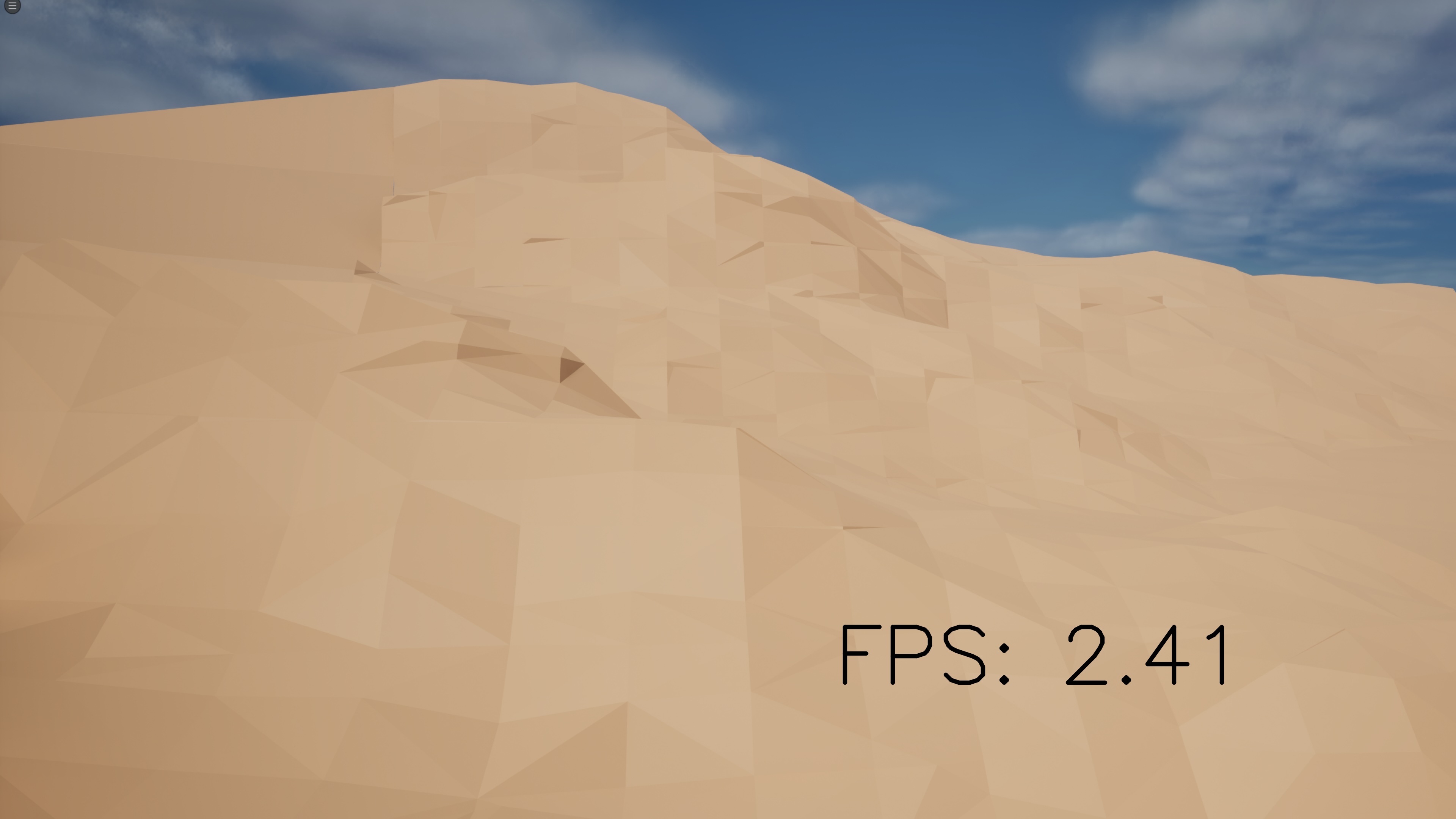}
   \caption{Re-extracting new view-dependent meshes with Infinigen's solution cannot achieve interactive frame-rates, even at extremely low detail ignoring any mesh load-time.}
   \label{fig:nonrealtime}
   \vspace{-4mm}
\end{figure}

Most importantly, our method can create a simulated virtual environment in real-time rendering engines like Unity or Unreal Engine, where an embodied AI agent can explore and learn. To demonstrate this, we took a scene from the previous dataset, and extracted a mesh that is valid for a range of camera rotations and translations along a path of interest. This results in an unbounded mesh that can be explored interactively within a certain range of camera poses. We export the resulting mesh to Unreal Engine. We experimented on a 4K ($3840\times 2160$) display and with one NVIDIA GeForce RTX 3090 GPU, and achieved \textgreater 50 FPS rendering frame rate. Fig.~\ref{fig:realtime}(a) shows the scene and the wireframe visualization, and we can see the mesh is denser on those surfaces closer to the camera view and sparser far away. Fig.~\ref{fig:realtime}(b) shows the screenshots of the real-time video. In contrast, \infinigen{} needs to create a new mesh for each new view, which takes about 0.5 seconds even in such low resolution as shown in  Fig.~\ref{fig:nonrealtime}. Even ignoring mesh loading overhead, this would give a theoretical framerate of 2.41 FPS. This could be increased by refreshing the mesh only every few frames, but this would produce visible seams and low poly faces.

\section{Conclusion}

We proposed a method to extract meshes for unbounded scenes based on a given SDF and a given set of cameras. Unlike previous methods, we generate a high-resolution mesh that can be reused for all predefined cameras. This is very useful in both generating view-consistent procedural synthetic datasets and providing a real-time virtual training environment for embodied computer vision AI.

\section{Acknowledgements}
This work was partially supported by the National Science Foundation and Amazon.

\newpage

{
    \small
    \bibliographystyle{ieeenat_fullname}
    \bibliography{main}
}

\clearpage
\setcounter{page}{1}

   {
   \newpage
       \onecolumn
        {
        \centering
        \Large
        \textbf{\thetitle}\\
        \vspace{0.5em}Supplementary Material \\
        \vspace{1.0em}
        }

   }

\section{More Quantitative Measurements of View Consistency}
\label{sec:supp1}
\begin{figure}[h]
  \centering
      \begin{subfigure}[t]{.19\linewidth}\centering
\includegraphics[width=\linewidth]{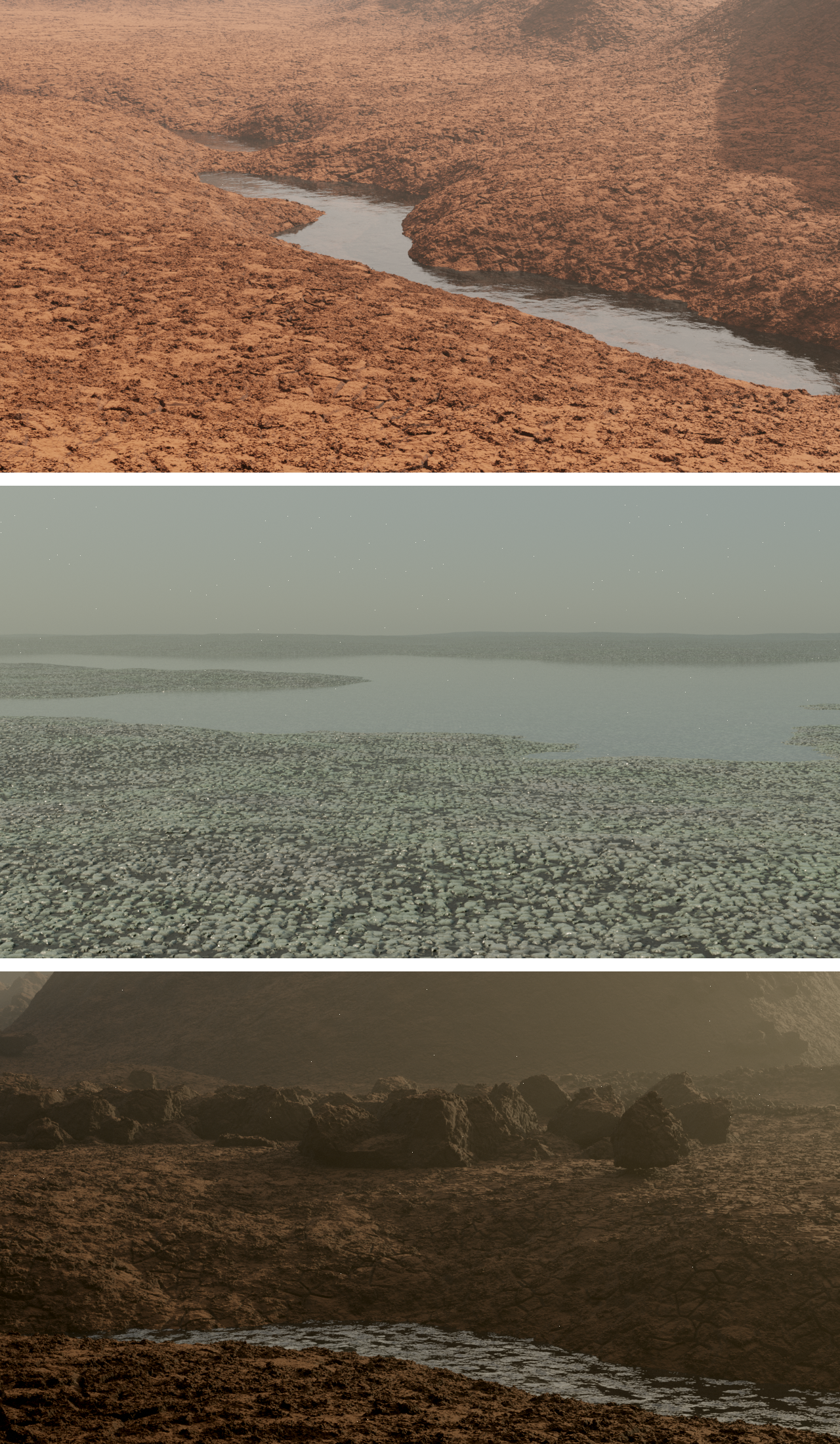}
    \caption*{(a) Scene overview}
    \end{subfigure}
        \begin{subfigure}[t]{.57\linewidth}\centering
\includegraphics[width=\linewidth]{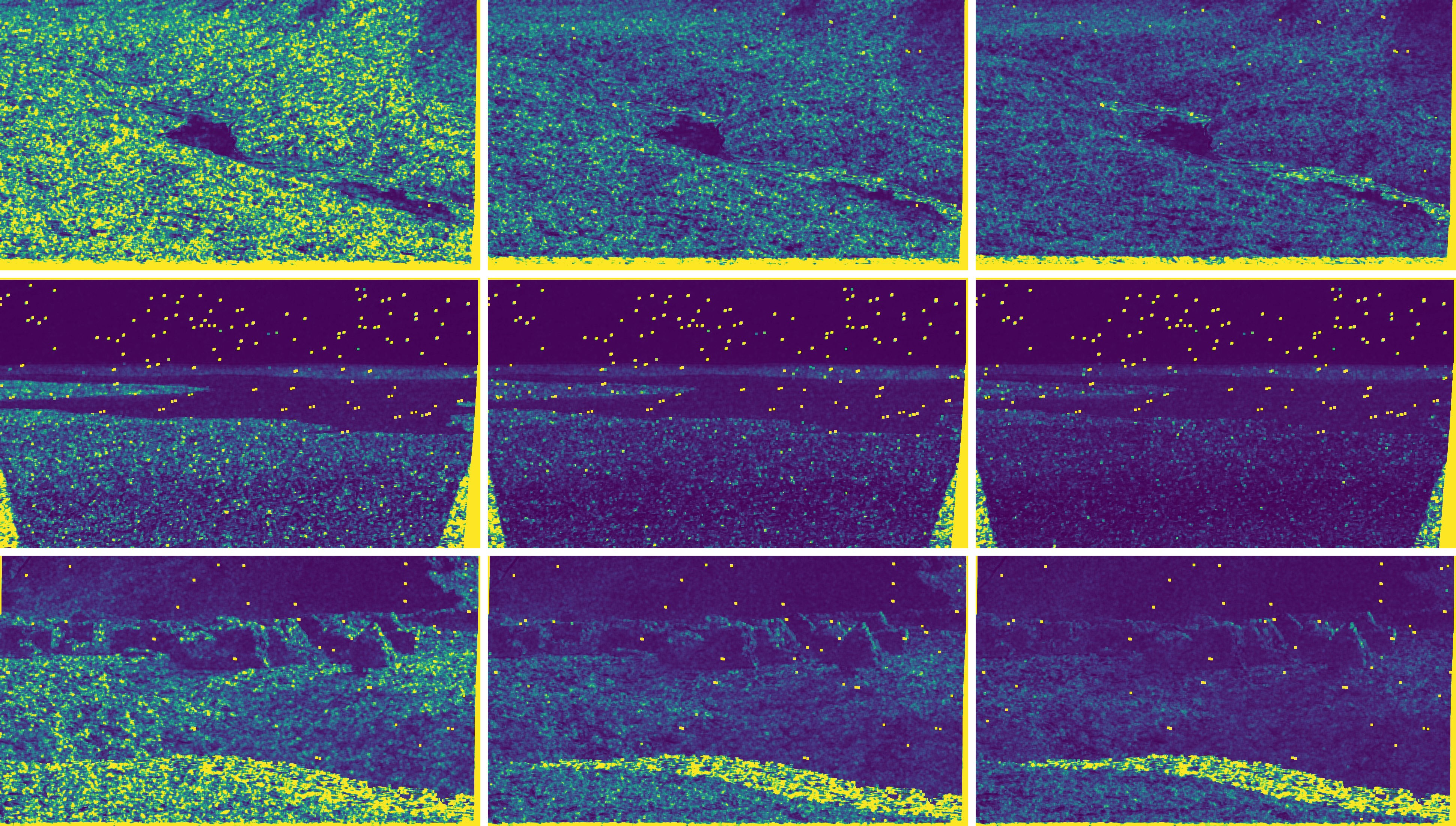}
    \caption*{(b) \infinigen{} with increasing resolution}
    \end{subfigure}
        \begin{subfigure}[t]{.19\linewidth}\centering
\includegraphics[width=\linewidth]{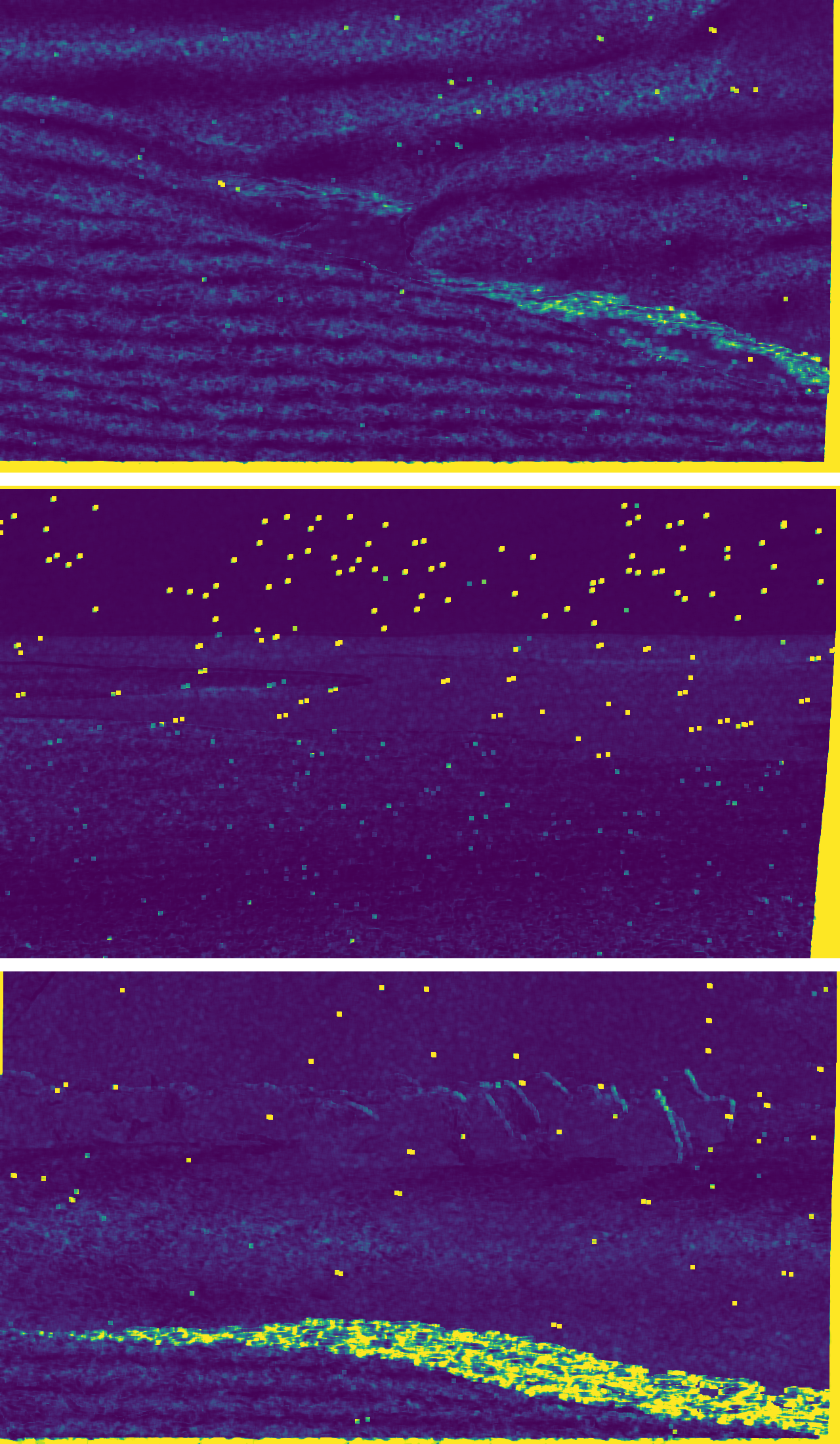}
    \caption*{(c) Ours}
    \end{subfigure}

   \caption{Quantitative measurement of view consistency for the first pair of transition frames. The brighter, the worse.}
\end{figure}

\begin{figure}[h]
  \centering
        \begin{subfigure}[t]{\linewidth}\centering
\includegraphics[width=0.46\linewidth]{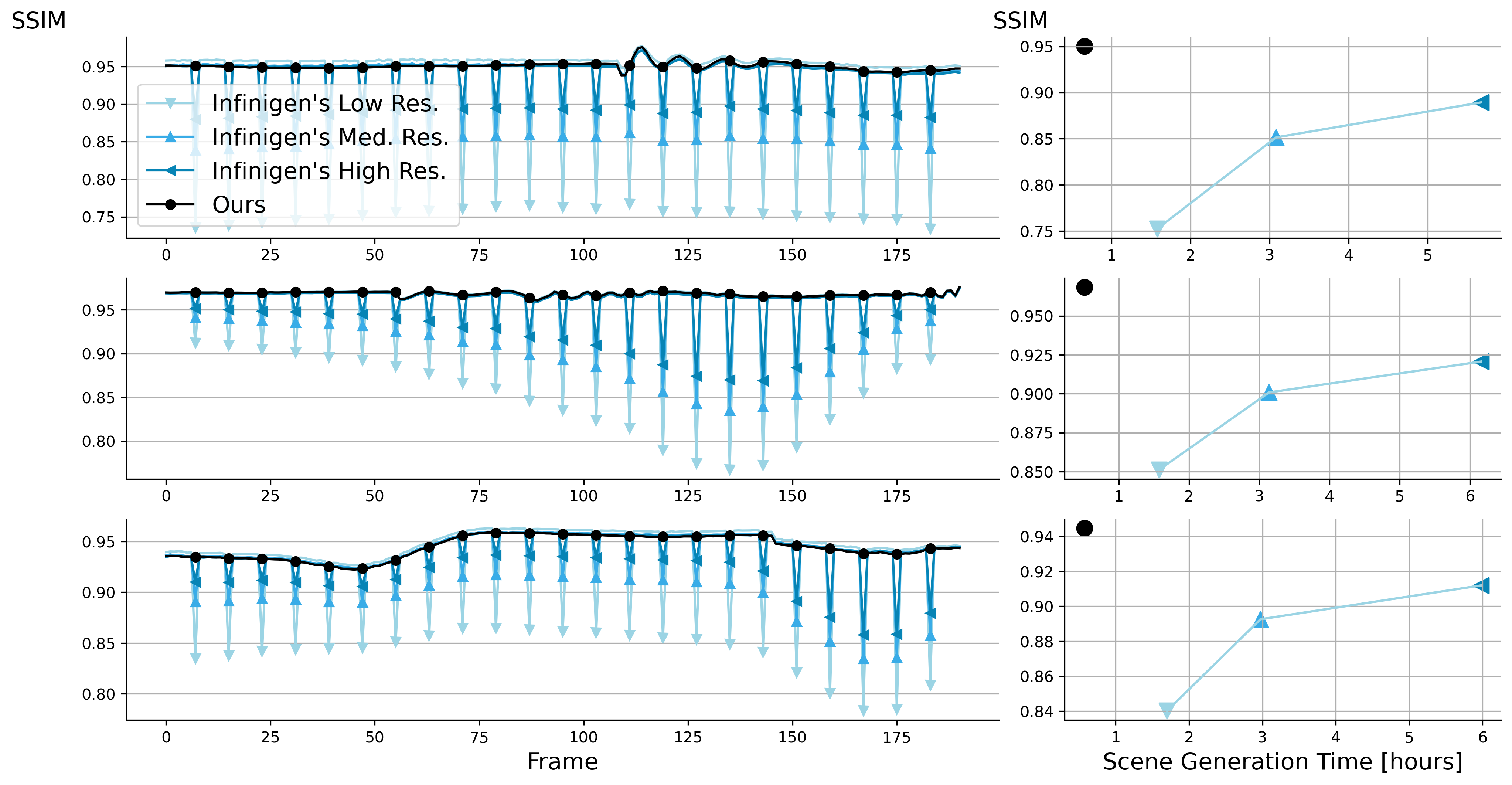}
    \caption*{(a) Comparison of frame-by-frame SSIM}
    \end{subfigure}
        \begin{subfigure}[t]{\linewidth}\centering
\includegraphics[width=0.46\linewidth]{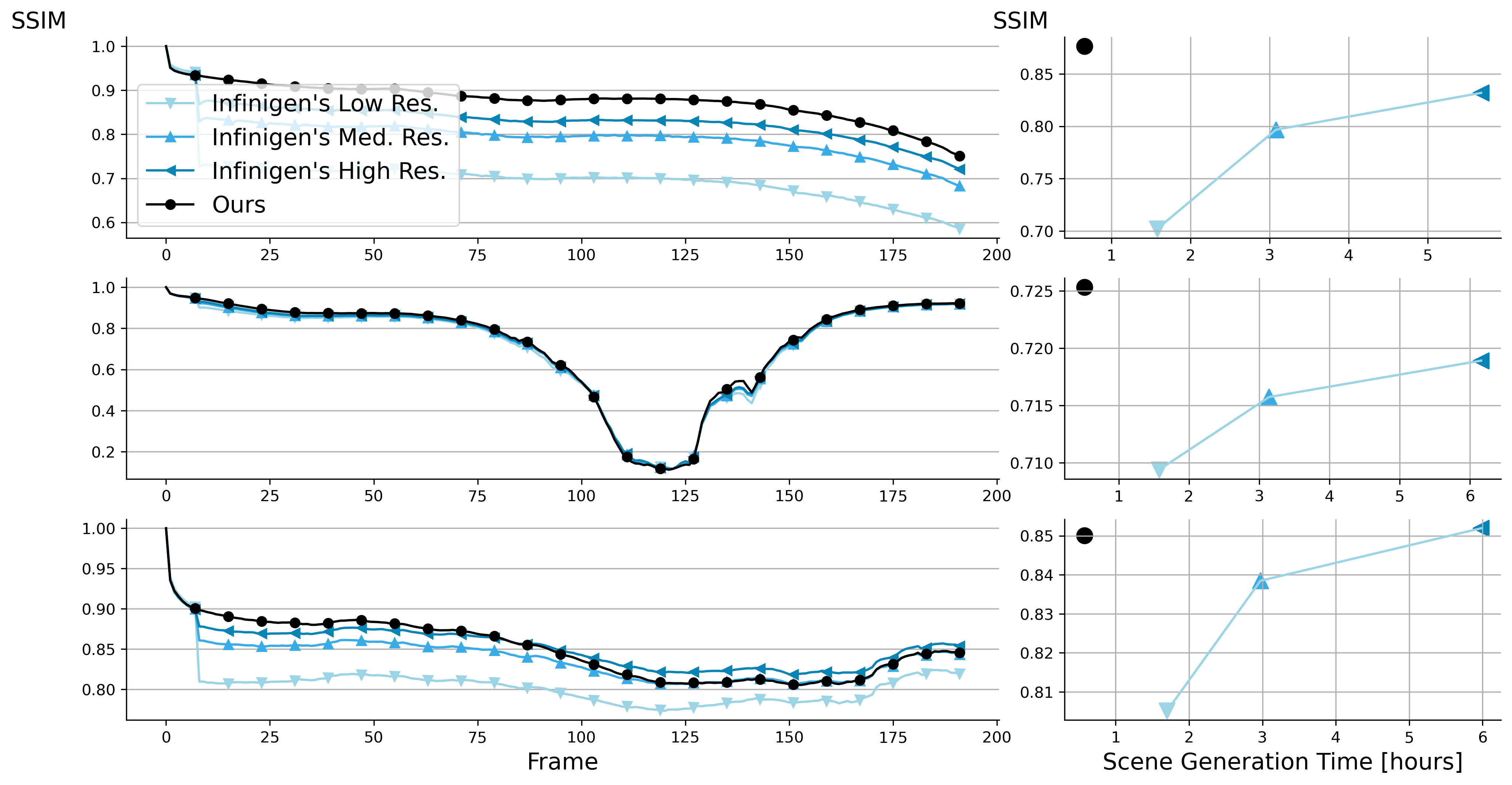}
    \caption*{(b) Comparison of first-to-nth frame SSIM}
    \end{subfigure}
    
   \caption{Quantitative measurement of view consistency (a) for adjacent frames and (b) along a pixel trajectory. Extension of Fig.~\ref{fig:ssim_tradeoff}}
   \vspace{-2mm}
\end{figure}

\section{More Evaluation Results of 3 Pre-trained Optical Flow Methods}
\label{sec:supp2}
\begin{figure}[h]
  \centering
        \begin{subfigure}[t]{0.22\linewidth}\centering
\includegraphics[width=\linewidth]{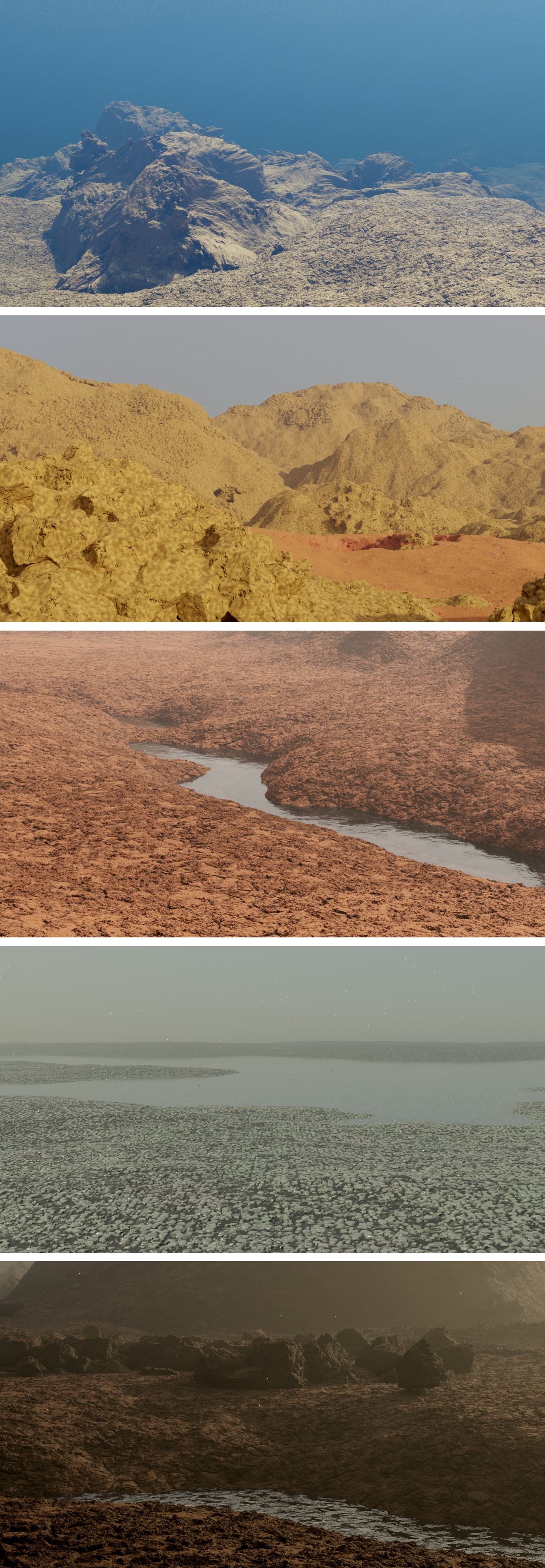}
    \caption*{(a) Scene overview}
    \end{subfigure}
        \begin{subfigure}[t]{0.76\linewidth}\centering
\includegraphics[width=\linewidth]{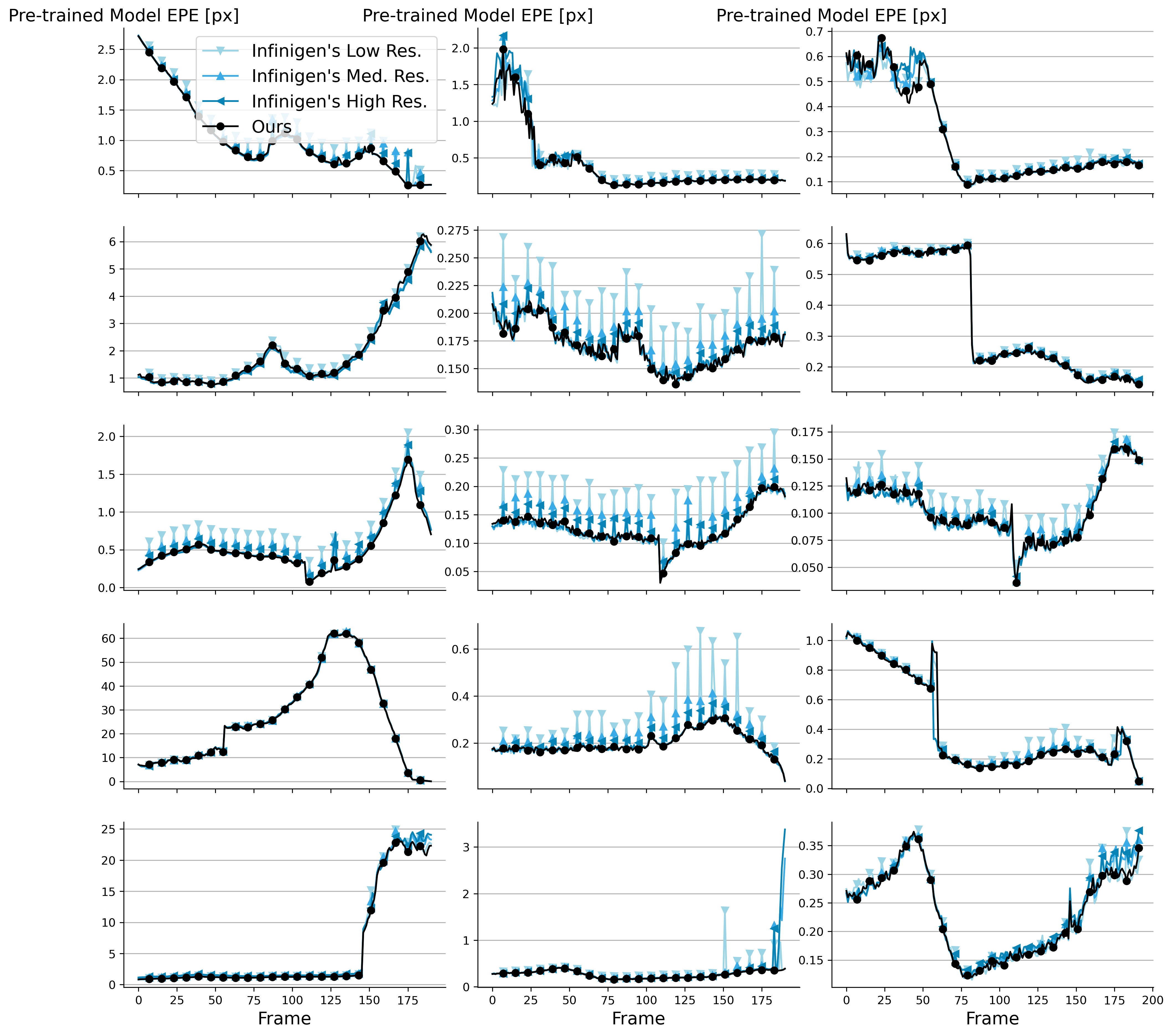}
    \caption*{(b) From left to right: Gunnar Farneback's algorithm, RAFT and VideoFlow}
    \end{subfigure}

   \caption{End-point-error (EPE) for 3 pre-trained optical flow methods evaluated on meshes from \projectname{} vs. Infinigen at various resolutions.}
   \vspace{-2mm}
\end{figure}

\end{document}

%% file: preamble.tex
\usepackage[dvipsnames]{xcolor}

\newcommand{\infinigen}{Infinigen}

\usepackage{multirow}

\usepackage{amssymb}%
\usepackage{pifont}%
\newcommand{\cmark}{\ding{51}}%
\newcommand{\xmark}{\ding{55}}%
\usepackage{amssymb}
\usepackage{multicol}